\pdfoutput=1

\documentclass[11pt]{article}

\usepackage[preprint]{acl}

\usepackage{times}
\usepackage{latexsym}

\usepackage[T1]{fontenc}

\usepackage[utf8]{inputenc}

\usepackage{microtype}

\usepackage{inconsolata}

\usepackage{xcolor}
\usepackage{amssymb}
\usepackage{amsmath}
\usepackage{amsfonts}
\usepackage{booktabs}
\usepackage{graphicx}
\usepackage{amssymb}
\usepackage{pifont}
\newcommand{\cmark}{\scalebox{1.5}{\textcolor{green}{\ding{51}}}}%
\newcommand{\xmark}{\scalebox{1.5}{\textcolor{red}{\ding{55}}}}%
\usepackage[most]{tcolorbox}

\usepackage{multirow}

\newcommand*\circled[1]{\tikz[baseline=(char.base)]{
            \node[shape=circle,draw,inner sep=0.3pt] (char) {#1};}}

\definecolor{agreen}{rgb}{0.55, 0.71, 0.0}
 \definecolor{lightcoral}{rgb}{0.94, 0.5, 0.5}
 
\newcommand{\vi}[1]{\textcolor{black}{#1}}

\newcommand{\inlineSubsection}[1]{
  \par\noindent\textbf{#1}\quad
}

\title{Evaluating and Improving Graph to Text Generation with Large Language Models }

\author{
    Jie He$^1$\footnotemark[1] \quad 
    Yijun Yang$^1$\thanks{\ \ Equal Contribution.} \quad 
    Wanqiu Long$^1$ \quad 
    Deyi Xiong$^2$ \quad \\ 
    \textbf{Victor Gutierrez Basulto}$^3$\footnotemark[2] \quad and \; 
    \textbf{Jeff Z. Pan}$^1$\thanks{ \, Corresponding author}  \\
     $^1$ School of Informatics, University of Edinburgh, UK  \\
     $^2$ College of Intelligence and Computing, Tianjin University, Tianjin, China\\
    $^3$ School of Computer Science and Informatics, Cardiff University, UK \\
    \normalsize{\texttt{ \{j.he, j.z.pan\}@ed.ac.uk, thomasyyj@outlook.com, }} \normalsize{\texttt{GutierrezBasultoV@cardiff.ac.uk}} \\
}

\begin{document}
\maketitle
\begin{abstract}
\textbf{L}arge \textbf{l}anguage \textbf{m}odels (LLMs) have demonstrated immense potential across various tasks. 
However, research for exploring and improving the capabilities of LLMs in interpreting graph structures remains limited.  To address this gap, we conduct a comprehensive evaluation of prompting current open-source LLMs on graph-to-text generation tasks. Although we explored the optimal prompting strategies and proposed a novel and effective diversity-difficulty-based few-shot sample selection method, we found that the improvements from tuning-free approaches were incremental, as LLMs struggle with planning on complex graphs, particularly those with a larger number of triplets.
To further improve LLMs in planning with graph sequences and grounding in truth, we introduce a new graph-to-text dataset, PlanGTG, annotated with two sub-tasks: reordering and attribution. Through extensive automatic and human evaluations, we demonstrate significant improvements in the quality of generated text from both few-shot learning and fine-tuning perspectives using the PlanGTG dataset. Our study paves the way for new research directions in graph-to-text generation\footnote{PlanGTG datasets can be found in \href{https://github.com/probe2/kg_text}{https://github.com/probe2/kg\_text.}.}

\end{abstract}

\section{Introduction}
Recent advancements in large language models  \cite{Chowdhery2022PaLMSL, openai2022chatgpt, openai2023gpt4, touvron2023llama, jiang2023mistral} have revolutionized \textbf{n}atural \textbf{l}anguage \textbf{p}rocessing (NLP) due to their remarkable zero- and few-shot capabilities. While LLMs have been explored for structured graph tasks \cite{Rong2020DropEdge} and graph classification \cite{Errica2020A}, their potential in verbalizing graphs in natural language (graph verbalization) remains underexplored. Graph-to-text  generation \cite{koncel-kedziorski-etal-2019-text, ribeiro-etal-2021-investigating} is a challenging task that yields text from different graph structures and requires semantic alignment between graph and text.


%

%


\textbf{K}nowledge \textbf{g}raphs (KGs) store graph-like knowledge in triplets  $\langle h, r, t \rangle$, stating that the head entity $h$ is related to the tail entity $t$ through the relation type $r$. Verbalizing triplets from KGs is essential for a wide range of  tasks, such as creating QA datasets from graph data, e.g.\ CommonsenseQA \cite{talmor-etal-2019-commonsenseqa} and SciGraphQA \cite{li2023scigraphqa}. It also plays a key role in mitigating hallucinations of LLMs \cite{agrawal2023knowledge, zhao-etal-2023-verify}. 

To advance the KG-to-text  generation task in the era of LLMs, we perform a preliminary evaluation on how well open-source LLMs perform on different prompts both in zero and few-shot scenarios. The prompt searching results emphasized the role of detailed instructions in unleashing LLMs' potential to generate fluent and accurate text from graphs. At the same time, we did not observe  improvements of LLMs over various prompt optimizations in the zero-shot case (e.g. different linearizations and triplet expressions). 
For few-shot prompts, we empirically showed that choosing moderately hard and diverse prompts results in the best performance and propose a novel difficulty-diversity balanced demonstration selection method, which outperforms both simple \textbf{d}ifficulty and \textbf{d}iversity-based \textbf{d}emos selection (DDD). However, the absolute value of improvements is marginal, reflecting the limitations of in-context learning. The analysis of graphs with varying complexity, measured by the number of triplets and graph diameters (the longest shortest path between any two vertices), further suggests that current LLMs struggle with planning when handling graphs with a lot of triplets and with small diameters. This motivates us to use instruction tuning over sub-tasks to strengthen LLMs' planning abilities, thus improving the performance of graph-to-text generation.




Inspired by \citet{zhao-etal-2023-structure}, we investigate whether instructing LLMs  to explicitly output their decision process for graph-to-text generation can improve the quality of  generated text. In addition, we design two subtasks: 1) \textbf{Reordering}: reorder the  given KG triplets to better align with the generated text. 2) \textbf{Attribution}:  attribute triplet indexes in the generated text using sequential numbers to enhance the interpretability of the generated text. 
%
To achieve this, we create   \textbf{PlanGTG} (Sec.\ref{sec6}), a new instruction dataset containing approximately 30,000 data pairs, featuring annotated attributions of triplets from the text. This dataset is  generated using seeds rewritten from GraphNarritive \cite{shi-etal-2023-hallucination} and GPT-3.5-turbo, incorporating the two subtasks sequentially.

We fine-tune LLMs with PlanGTG (Sec.\ref{sec7}), conducting extensive automatic and human evaluations. Human evaluation results suggest that models fine-tuned with instructions   are capable of successfully adjusting the order of KG triplets and correctly marking the sequence numbers in the generated text in most cases. Comparatively, models trained with PlanGTG outperform those using the EventNarrative \cite{colas2022eventnarrative}, TEKGEN \cite{agarwal-etal-2021-knowledge}, and GraphNarrative \cite{shi-etal-2023-hallucination} datasets in both zero-shot generalization and full-shot fine-tuning.

\vi{In summary, our main contributions are:}
\begin{itemize}
\setlength{\itemsep}{0pt}
\setlength{\parsep}{0pt}
\setlength{\parskip}{0pt}

\item{We conduct comprehensive preliminary evaluations of graph-to-text tasks on LLMs, and explore the most effective prompting strategies in zero-shot and few-shot cases. We propose a novel and effective demonstration selection method DDD, and point out the remaining challenge of planning on complex graphs.}
\item{We construct the PlanGTG dataset by adding two new subtasks, reordering and attribution, 
to study how LLMs can be improved through instruction fine-tuning with enriched auxiliary task information.}
\item{Extensive experiments and evaluations have validated the effectiveness and utility of PlanGTG. Additionally, we explored how a curriculum learning approach, which strategically organizes the sequence of training data, can further enhance model performance.}
\end{itemize}

\section{Related Work}
Various approaches have been proposed for transforming knowledge graphs  into text. These include graph neural network based methods \citep{ribeiro2020modeling} and language model based approaches \citep{liu-etal-2022-syntax, zhao-etal-2023-structure}. Graph neural network based methods typically encode structured inputs explicitly as model representations \cite{Puduppully_Dong_Lapata_2019, 10.1162/tacl_a_00269, koncel-kedziorski-etal-2019-text}. 
Among language model based approaches, a critical step is linearizing input triplets \citep{zhao-etal-2020-bridging}, and recent efforts have introduced various planning techniques \citep{zhao-etal-2020-bridging, zhao-etal-2023-structure}. 

There are  studies that  have focused on evaluating graph-to-text generation quality using pre-trained language models. \citet{shi-etal-2023-hallucination} address hallucinations in open-domain graph-to-text generation, while \citet{yuan2023evaluating} evaluate closed-source LLMs such as GPT-3 and ChatGPT only under the  zero-shot setting. In contrast, our work focuses on open-source LLMs, aiming to provide a comprehensive evaluation across various aspects of the graph-to-text  conversion process, including linearization, demonstration selection, and model scaling. Furthermore, we propose two new sub-tasks, reordering and attribution. These tasks are designed to enhance transparency in the generation process of LLMs and improve their performance.

\section{Preliminary Study}
To investigate how to improve the performance of LLM's graph-to-text generation, we first perform a preliminary study to answer 1) How can different prompts influence the performance of graph-to-text generation, and will prompt engineering be relevant to attain such improvements? 2) Will scaling the parameters of open-sourced LLMs improve their graph-to-text verbalization abilities? We performed experiments following the settings in Section \ref{exp_setup}. For detailed experiment results, please refer to App. \ref{appendix:evaluation}.

\subsection{\vi{Experimental Setup}}
\label{exp_setup}
\inlineSubsection{\vi{Datasets}} 
\vi{We conducted all of our experiments on the following three benchmarks: \textbf{WebNLG17}, \textbf{WebNLG20},
\textbf{DART}. Details about the datasets and  experimental setup can be found in App.~\ref{app:data}}.

\label{human_eval}
\inlineSubsection{\vi{Evaluation Metrics}} We used both automatic evaluations and human evaluations for our experiments. Four popular metrics BLEU (B-4) \cite{papineni-etal-2002-bleu}, METEOR (ME) \cite{banerjee-lavie-2005-meteor}, CHRF++ (CF) \cite{popovic-2015-chrf} and BartScore (BS) \cite{yuan2021bartscore} are adopted as evaluation metrics.  Detailed explanations for all metrics are given in App. \ref{appendix:evaluation_metric}. 
\inlineSubsection{Model} All preliminary experiments except the scaling study are performed on Mistral-7b-instruct-v0.2 \cite{jiang2023mistral}.

\subsection{Prompt Searching}
\label{sec:prompt_searching}
We first explored how different instructions and the textual representation of triplets influence downstream performance in zero-shot cases. \emph{This allows us to fix the optimal prompts as the prompt format for the rest of the paper.} In summary, more detailed prompts for explaining the formulation of the provided triplets and the task's goal result in more accurate and coherent generations. Regarding triplet formats,  ``\textit{head | relation | tail}'' performs slightly better than other representations such as ``$\langle \textit{head} \rangle$ $\langle \textit{relation} \rangle$ $\langle \textit{tails} \rangle$'', but not significantly. Similarly, although several works have demonstrated the significance of different linearization techniques in graph-to-text tasks \cite{yang-etal-2020-improving-text, hoyle-etal-2021-promoting, li-etal-2021-shot-knowledge}, we found that LLMs are robust over different linearization and we, therefore, applied the default linearizations provided in the datasets in the rest of experiments. We conjecture that this robustness is due to LLMs being more invariant with the position and order, likely resulting from updated position embeddings such as RoPE \cite{su2024roformer}. Detailed experiments and discussions are available in App. \ref{appendix:instruction_explore}.
\subsection{Example Selection}
As the prompt templates are fixed during the prompt searching, we explored how few-shot demonstrations selection can influence and improve the performance of graph-to-text generation motivated by research emphasizing the importance of demonstrations selection \cite{luo2023dr, drozdov2023parade}. We conducted experiments on selecting demonstrations based on difficulty and diversity. Here difficulty is assessed by computing the cosine similarity between examples and input, with the assumption that following similar demonstrations facilitates easier graph-to-text generation. Diversity is measured by whether examples are within different $k$-means clusters. We will summarize our main findings due to the page limits. Please refer to App. \ref{appendix:example_selection} for detailed experiment results. 
\begin{table}[!tph]
\centering\small
\setlength{\tabcolsep}{0.6ex}
\begin{tabular}{lccccc}
\toprule
  \multicolumn{1}{l}{Dataset} &
  \multicolumn{1}{l}{\# of shots} &
  \multicolumn{1}{c}{B-4} &
  \multicolumn{1}{c}{ME} &
  \multicolumn{1}{c}{CF} &
  \multicolumn{1}{c}{BS} \\
  \midrule
  \multirow{4}{*}{DART} &
  \phantom{0}1 & 21.33 & 33.69 & 55.10 & -2.36 \\ 
&\phantom{0}3  & 21.30 & 33.78 & 55.17 & \textbf{-2.33} \\
&\phantom{0}5  & 22.22 & 33.90 & 55.47 & -2.35 \\
&10            & \textbf{23.32} & \textbf{34.20} & \textbf{56.00} & -2.35 \\
\midrule
  \multirow{4}{*}{WebNLG17} &
  \phantom{0}1& 23.48 & 35.49 & 57.66 & -1.92 \\ 
&\phantom{0}3 & 25.07 & \textbf{35.98} & 58.60 & \textbf{-1.91} \\
&\phantom{0}5 & 24.83 & 35.92 & 58.72 & -1.94 \\
&10           & \textbf{25.28} & 35.87 & \textbf{58.85} & -1.96 \\
\midrule
  \multirow{5}{*}{WebNLG20} &
  \phantom{0}1& 26.07 & \textbf{38.01} & 58.64 & \textbf{-1.97} \\ 
&\phantom{0}3 & \textbf{27.54} & 36.09 & \textbf{58.93} & -2.00 \\
&\phantom{0}5 & 27.42 & 35.95 & 58.58 & -2.03 \\
&10           & 27.36 & 36.01 & 58.87 & -2.03 \\
\bottomrule
\end{tabular}
\setlength{\belowcaptionskip}{-0.4cm}
\caption{Comparison between different numbers of few-shot selections on three popular graph-to-text datasets. B-4, ME, CF, BS refer to BLEU-4, METEOR, CHRF++ and BartScore respectively}
\label{shot_numbers}
\end{table}

\paragraph{\textit{Finding 1:} Selecting moderately difficult and diverse demonstrations yields the best results, but with fluctuations.}
We separate the level of difficulties into 5 levels by uniformly selecting examples ranging from the most similar to the least similar examples between the input and training sets considering both graph similarity and generated text similarity in the one-shot scenario. For the performance shown in Fig. \ref{Difficulty_full} and Fig. \ref{Diversity_full}, we found that in most cases, the performance peaks at the middle level of both difficulty and diversity. This suggests that LLMs generalize better due to not relying on memorized shortcuts from overly similar demonstrations or samples closely resembling the input. Although these findings give insights on the demonstrations selection and motivate us to propose the \textbf{d}ifficulty-\textbf{d}iversity balanced \textbf{d}emos selection (DDD), there are cases where results fluctuate unpredictably based on different demonstrations. We attribute this variability to potential limitations in the engineering of demonstrations.
\paragraph{\textit{Finding 2:} DDD selections perform better than simple diversity and difficulty selection, but marginally. }
\vi{Based on our previous findings, we propose a DDD selection approach. This method employs a dual-phase process where initially, leveraging findings from our diversity investigation, we recall all samples within the same cluster. These samples are then sorted by their difficulties (i.e.\ the cosine similarity from the input graph). 
For instance, in a 3-shot scenario, we select samples ranked at  25\%, 50\%, and 75\%  in terms of similarity. The comparative analysis presented in Table \ref{DDD} shows that DDD is more effective across various datasets than strategies solely focused on difficulty or diversity. This highlights that concurrently considering difficulty and diversity in the selection of demonstration samples could be the best sample selection strategy. However, despite the good comparative results, the absolute value of improvement is incremental both from the best result of one-shot to three-shot and from DDD selection to difficulty/diversity selection. This suggests that few-shot example selections yield minimal improvements. }

\paragraph{\textit{Finding 3:} Increasing the number of shots does not help.} To mitigate the high randomness of sampling demonstrations, we investigate the influence of increasing the number of demonstrations based on the DDD selection methods. We picked 1, 3, 5, and 10 samples and reported the results in Table \ref{shot_numbers}. We found that increasing the number of samples does not always help. This can possibly explained by the fact that the samples selected from DDD methods are already diverse enough for LLMs to learn the task format and those difficult demonstrations did not provide additional relevant knowledge to help LLMs translate the input graph.

\begin{table}[]
\centering\small
\setlength{\tabcolsep}{1ex}
\begin{tabular}{lccccc}
\toprule
  \multicolumn{1}{c}{Number} &
  \multicolumn{1}{c}{(\#samples)} &
  \multicolumn{1}{c}{B-4} &
  \multicolumn{1}{c}{ME} &
  \multicolumn{1}{c}{CF} &
  \multicolumn{1}{c}{BS} \\
  \midrule
  \multirow{5}{*}{triplets } &
  1\phantom{000} (848) & 23.36 & 38.93 & 61.22 & -1.92 \\ 
&2\phantom{000} (797) & 17.55 & 32.98 & 53.59 & -2.36 \\
&3\phantom{000} (821) & 17.09 & 32.57 & 53.24 & -2.22 \\
&4\phantom{000} (869) & 17.21 & 31.68 & 52.89 & -2.29 \\
&5+\phantom{0} (1762) & 11.79 & 29.59 & 47.08 & -2.63 \\
\midrule
  \multirow{5}{*}{Diameters } &
  0\phantom{000} (575)& 23.29 & 38.92 & 61.19 & -1.93 \\
&1\phantom{00} (3463)& 14.18 & 30.30 & 48.65 & -2.53 \\
&2\phantom{000} (850)& 18.60 & 33.20 & 55.09 & -2.00 \\
&3\phantom{000} (201)& 17.75 & 32.37 & 54.41 & -2.02 \\
&4\phantom{00000} (8)& 20.57 & 33.96 & 59.36 & -2.24 \\
\bottomrule
\end{tabular}
\setlength{\belowcaptionskip}{-0.4cm}
\caption{Comparison of performance between different triplet numbers and graph diameters on DART dataset.}
\label{improve_space}
\end{table}
\begin{figure*}[t]
\centering
\includegraphics[width=1\linewidth,height=0.8\linewidth]{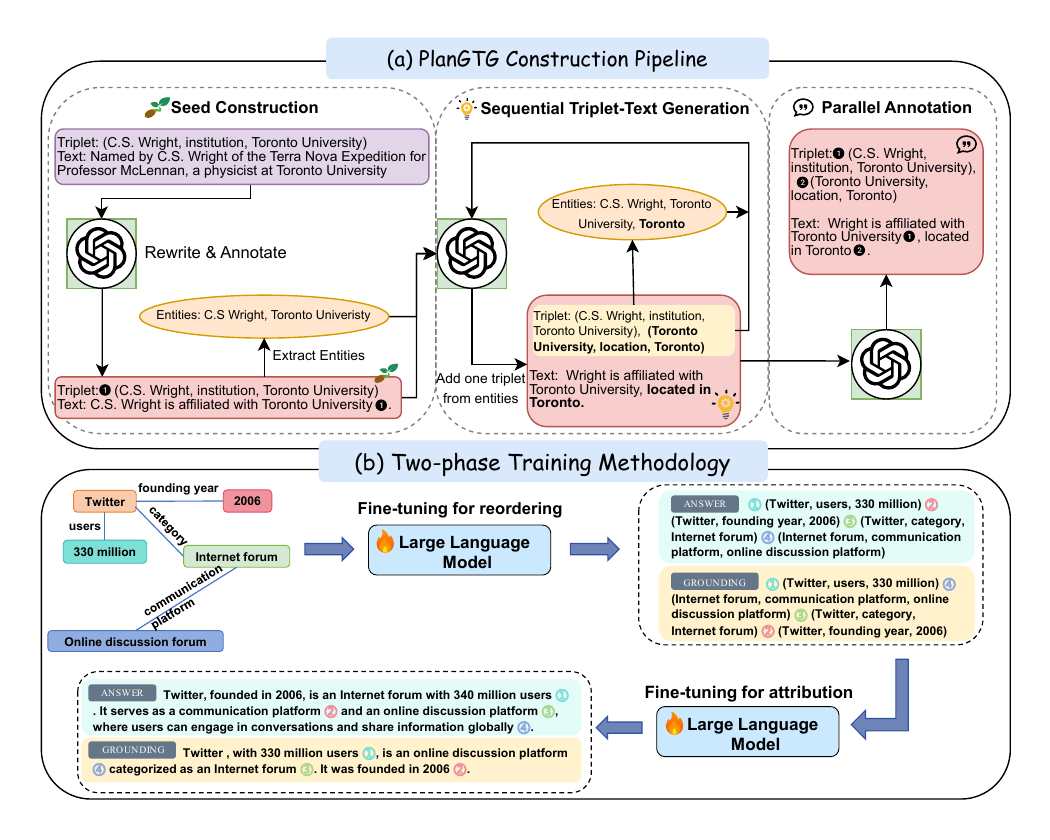}
\setlength{\abovecaptionskip}{-0.6cm}
\setlength{\belowcaptionskip}{-0.4cm}
\caption{(a) The flow chart of the construction for the PlanGTG dataset, the squared text refers to the output of GPTs and the circled text represents the result extracted automatically by rules. The newly added information by GPTs is marked in bold. (b) Our training pipeline: The training methodology consists of two phases: planning-guided generation and attribution generation. It enables LLMs to first generate triplets that follow a more natural language order and subsequently guide the generation of attributed answers.}
\label{pipeline}
\end{figure*}

\subsection{Improvment Space}
Finally, we analyze the behavior of models over different complexity of graphs to explore the space for improvements. We represent the graph complexity through the number of triplets and the graph diameters (the longest-shortest path between any two vertices) and analyze the results on Mistral-7b-instruct-v0.2 in Table \ref{improve_space}\footnote{For results on all datasets, see Table \ref{graph_complex} and Table \ref{graph_diameter} in App. \ref{Appendix: GraphComplexity}.}. Firstly, as the number of triplets increases, a consistent decline in performance is shown across all metrics. This shows the difficulties that LLMs face in text generation from complex graphs. This motivates us to consider improving the planning abilities of LLM to handle graphs with more triplets. For graph diameters, LLMs perform the worst when interpreting graphs with only one diameter. This is because when the number of triplets is small, LLMs may suffer from hallucinations in order to make the generation coherent and satisfying-looking. This further motivated us to design a dataset that allows LLMs to cite their generation in order to mitigate the hallucination.

\section{The PlanGTG Dataset}
\label{sec6}
In the perspective of fine-tuning for improving the performance of graph-to-text generation, we construct PlanGTG (\textbf{Plan}ning for \textbf{G}raph-to-\textbf{T}ext \textbf{G}eneration), a graph-to-text paired instruction tuning dataset, annotated for both reordering and attribution subtasks. The construction of PlanGTG is  guided by several key objectives: (1) ensuring the diversity of both the structure of graphs (size and diameter) and the topic of texts; (2) avoiding text-graph misalignment (i.e. textual descriptions containing information not found in input graphs), a significant cause of hallucination \cite{shi-etal-2023-hallucination}; and (3) ensuring interpretability by annotating the attribution triplets from the text descriptions and automatically formatting the linearization labels. These goals are achieved through sequential generation and parallel annotation.

\subsection{Dataset Construction}
The flow chart for the process of generating PlanGTG is shown in (a) of Fig. \ref{pipeline}, which consists of three parts: seed data preparation, sequential graph-text pair generation and parallel attribution annotations. We apply GPT-3-turbo-1106 as the base model for data generation. For seed preparation, sequential pairs generation and the parallel annotation, the used prompts are shown in Fig. \ref{PlanGTG_prompt_rewrite}, Fig. \ref{PlanGTG_prompt_sequential} and Fig. \ref{PlanGTG_prompt_parallel} in Appendix respectively.
\inlineSubsection{\vi{Seed Data Creation}}
\vi{We begin with the graph-text pairs in GraphNarritive \cite{shi-etal-2023-hallucination}. As a start,  we choose data-text pairs with only one triplet. To ensure diversity, we uniformly selected $n$ random samples from each type of relation present in GraphNarritive. Next, we prompt GPT-3.5-turbo-1106 to `regenerate' the text based on the source text and graphs, incorporating the annotation derived from the graph. For instance, given the triplet ``\textit{C.S. Wright | institution | Toronto University}'' alongside the text ``\textit{named by C.S. Wright of the Terra Nova Expedition for Professor McLennan, a physicist at Toronto University}'', we use GPT to regenerate the text as ``\textit{C.S. Wright is affiliated with Toronto University. (1)}''. This allows us to make both the initial attribution annotation and discard the redundant information from the initial description text. We create about 3 thousand one-triplet seeds to serve as the foundation data. }


\paragraph{\vi{PlanGTG Generation}}
\label{sec4.2}
\vi{We then generate the PlanGTG dataset from the foundation data in a sequential way. Specifically, for each triplet-text pair, we ask ChatGPT to (1)  generate the graph by adding one new triplet that integrates well with the existing triplets, (2) then update the corresponding text description incorporating information of the added triplet and (3) make the attribution annotation. We conduct one inference for Steps 1 and 2 and perform a separate inference for Step 3 and different demonstrations are provided for Step 3. This is because we empirically found that integrating the attribution annotation numbers may harm the performance of generating the text description of the new triplets. We also provide the existing entities from the triplet list and guide GPT to choose one of the entities in Step 1 to ensure the connectivity of the generated graph. The above process shows the steps to craft one sample with $n+1$  triplets from $n$ triplets. For each foundation seed, we do these steps iteratively to attain 2 to 10 triplets. Eventually, we created PlanGTG with 28,837 training points and 996 development points after filtering data points with wrong patterns.}

\paragraph{\vi{Dataset Description}}In general, PlanGTG consists of 28,837 training pairs and 996 development graph-text pairs, with an average of 5.48 triplets within graphs and 37.6 words in the text. In the Appendix \ref{Appendix: PlanGTGDescription}, Fig \ref{PlanGTG_graph_triplet} shows the distribution of the number of triplets in the dataset. Fig \ref{PlanGTG_graph_dia} shows the distribution of the diameters in the graph. Fig \ref{PlanGTG_text} shows the distribution of the words contained in the text. To ensure that PlanGTG does not overlap with the test sets, we checked all triplets in PlanGTG. The percentages of overlapping triplets across datasets are 0\% for DART, 0\% for WebNLG 2017, and 0.00019\% for WebNLG 2020 test sets. Additionally, none of the input graphs overlap with these test sets.

\subsection{\vi{Dataset Quality} }
\begin{table}[]
\setlength{\tabcolsep}{0.9ex}
\tiny
    \centering
    \begin{tabular}{p{25pt}|p{35pt}p{20pt}p{25pt}p{25pt}|p{10pt}} 
    \toprule
    Dataset&Hallucinated Entities$\downarrow$
&Missed Entities$\downarrow$ &Hallucinated Relations$\downarrow$ &Missed Relations$\downarrow$ &GR$\uparrow$  \\
    \midrule 
    TEKGEN&0.84&0.08&0.92&0.07&4.48\\
    EVENT&0.69&0.05&0.70&0.08&4.69\\
    GN&0.62&0.02&0.74&0.03&4.73\\ 
    PlanGTG&0.36&0.01&0.42&0.03&4.78\\
    \bottomrule
    \end{tabular}
    \setlength{\belowcaptionskip}{-0.2cm}
    \caption{Human evaluation on pretraining datasets. Co
hen’s kappa coefficients for labeling three factors are
 as follows: 0.82, 0.79, and 0.77. GN represents the GraphNarrative dataset. GR means the Grammar. }
    \label{6.1}
    \vspace{-0.3cm}
\end{table}

\begin{table*}[!ht]
\scriptsize
    \centering
    \begin{tabular}{l|l|cccc|cccc|cccc}
    \toprule
        \multirow{2}{*}{\textbf{Model}}&\textbf{Dataset}&\multicolumn{4}{c}{WebNLG17}&\multicolumn{4}{c}{WebNLG20}&\multicolumn{4}{c}{DART}  \\ 
        \cmidrule(lr){3-6}\cmidrule(lr){7-10}\cmidrule(lr){11-14}
        &\textbf{\#Metrics}&B-4& ME&CF&BS&B-4& ME&CF&BS&B-4& ME&CF&BS \\ \midrule
       \multirow{5}{*}{LLaMA2-7b-chat}& Zero & 17.32 & 28.11 & 45.82 & -2.85 & 17.45 & 23.62 & 38.94 & -2.81 & 14.14 & 29.29 & 46.58 & -2.85 \\ 
        &EVENT & 5.26 & 20.93 & 32.03 & -3.32 & 5.83 & 21.37 & 30.89 & -3.67 & 5.47 & 20.48 & 32.13 & -3.49 \\ 
        &GN & 9.99 & 22.78 & 36.05 & -3.14 & 9.04 & 20.76 & 33.14 & -3.35 & 12.33 & 22.63 & 39.48 & -3.55 \\ 
        &TEKGEN & 7.21 & 15.64 & 27.06 & -3.96 & 4.74 & 14.22 & 24.89 & -3.78 & 12.04 & 22.06 & 33.76 & -3.70 \\ 
        &Ours & \textbf{28.76} & \textbf{30.88} & \textbf{50.80} & \textbf{-2.38} & \textbf{20.60} & \textbf{25.50} & \textbf{42.16} & \textbf{-2.51} & \textbf{21.44} & \textbf{31.58} & \textbf{51.30} & \textbf{-2.72} \\ \midrule
         \multirow{5}{*}{Mistral-7b-chat} &Zero & 16.60 & 26.14 & 45.98 & -2.67 & 17.42 & 23.07 & 37.50 & -2.74 & 15.25 & 26.23 & 47.71 & -3.07 \\ 
        &EVENT & 6.48 & 23.06 & 34.86 & -3.01 & 6.34 & 22.34 & 33.71 & -2.99 & 7.87 & 23.50 & 37.70 & -3.48 \\ 
        &GN & 10.44 & 23.23 & 35.27 & -3.08 & 10.20 & 20.98 & 31.75 & -3.08 & 17.44 & 24.43 & 41.95 & -3.22 \\ 
        &TEKGEN & 8.18 & 21.46 & 32.81 & -2.82 & 4.92 & 18.86 & 29.11 & -2.81 & 13.44 & 25.52 & 37.93 & -3.10 \\ 
        &Ours & \textbf{29.76} & \textbf{31.30} & \textbf{51.17} & \textbf{-2.36} & \textbf{19.95} & \textbf{25.79} & \textbf{42.06} & \textbf{-2.43} & \textbf{27.46} & \textbf{30.50} & \textbf{50.21} & \textbf{-2.69} \\ \bottomrule
    \end{tabular}
    \setlength{\abovecaptionskip}{0.2cm}
    \setlength{\belowcaptionskip}{-0.1cm}

    \caption{Zero-shot performance of different methods for graph-to-text generation on three
 domains.  B-4, ME, CF and BS are short for BLEU-4, Meteor, CHRF++ and Bartscore. GN is short for GraphNarrative. }
    \label{6.2.1}

\end{table*}
\begin{table*}[!ht]
\setlength{\tabcolsep}{0.9ex}
\scriptsize
    \centering
    \begin{tabular}{lcccccccccccc}
    \toprule
         \textbf{\#Metrics}&  \multicolumn{3}{c}{B-4}   & \multicolumn{3}{c}{ME} &  \multicolumn{3}{c}{CF} & \multicolumn{3}{c}{BS}    \\ 
        \cmidrule(lr){2-4}\cmidrule(lr){5-7}\cmidrule(lr){8-10}\cmidrule(lr){11-13}
\textbf{Methods}&All&Seen&Unseen&All&Seen&Unseen&All&Seen&Unseen&All&Seen&Unseen\\        \midrule
        Direct FT & 45.32 & 49.83 & 39.53 & 35.47 & 37.50 & 33.06 & 62.04 & 65.16 & 58.18 & -2.17 & -2.06 & -2.29  \\ 
        Instruction FT & 36.34 & 40.09 & 31.54 & 33.39 & 34.89 & 31.62 & 56.10 & 58.61 & 53.01 & -2.54 & -2.48 & -2.61  \\ \midrule
        EVENT & 36.76 & 39.16 & 33.69 & 33.39 & 33.72 & 32.99 & 56.36 & 57.38 & 55.13 & -2.4 & -2.46 & -2.34  \\ 
        GN & 43.18 & 47.58 & 37.56 & 36.86 & 38.64 & 34.76 & 61.95 & 65.12 & 58.03 & -2.38 & -2.28 & -2.40  \\ 
        TEKGEN & 44.12 & 48.40 & 38.68 & 37.02 & 38.76 & 34.95 & 62.55 & 65.50 & 58.93 & -2.30 & -2.14  & -2.38  \\ \midrule
        Ours & \textbf{46.35} & \textbf{50.40} & \textbf{41.17} & \textbf{37.94} & \textbf{39.71} & \textbf{35.84} & \textbf{63.92} & \textbf{66.95} & \textbf{60.18} & \textbf{-1.91} & \textbf{-1.83} & \textbf{-2.00}  \\ \bottomrule
    \end{tabular}
    \setlength{\abovecaptionskip}{0.2cm}
    \setlength{\belowcaptionskip}{-0.2cm}
    \caption{Performance  of LLaMA2-7b-chat on WebNLG17 test set when fine-tuned with EVENT, GraphNarrative, TEKGEN and further
 fine-tuned with WebNLG17. Direct FT denotes that we directly fine-tune the model on WebNLG17 without adding instructions. Instruction  FT adopts the same instructions as the second instruction in Appendix \ref{ft_ex}.}
    \label{full_ft}    
\end{table*}
Two professional human annotators assess the quality of the generated graph-text pairs in PlanGTG.  We randomly select 200 examples and the annotators evaluate the hallucination, missing information, grammatical correctness and fluency of the generated text (using a 5-point Likert \cite{zis-Likert1932A}). Detailed explanations for all metrics are given in App. \ref{appendix:evaluation_metric}. The scores from both annotators are averaged. At the same time, we also evaluate the quality of automatically extracted pre-training texts (EVENT, TEKGEN, and GraphNarrative) in the same way. Table \ref{6.1} reveals that, on average, there are 0.36 hallucinated entities and 0.42 hallucinated relations per graph-text pair in PlanGTG. This reflects that our instruction dataset, although generated by ChatGPT, maintains high quality due to carefully designed generation instructions and meticulous post-processing. This quality is notably superior to automatically extracted pre-training texts, with GraphNarrative exhibiting the highest indicators of 0.62 for hallucinated entities and 0.74 for hallucinated relations.

In contrast, our evaluation shows low rates of missing entities and relations at 0.01 and 0.03, respectively, indicating that ChatGPT consistently incorporates graph information into text without significant information loss. Regarding language naturalness, the score 4.84 demonstrates that the generated text is highly fluent with minimal grammatical errors. These results highlight the high quality of our generated text compared to automatically extracted pre-training texts. For instance, GraphNarrative's best scores for missing entities and relations are 0.02 and 0.03, respectively, and its language naturalness score is 4.73, indicating slightly lower performance in these aspects.

\section{Experiments}
\label{sec7}
\subsection{\vi{Experimental setting}} \vi{To evaluate whether PlanGTG can enhance LLMs  generalizability in graph-to-text tasks, we conducted experiments on both zero- and full-shot  learning. 
In our zero-shot experiments, as shown in (b) of Figure \ref{pipeline}, we fine-tuned LLMs on the PlanGTG dataset and subsequently evaluated their performance on WebNLG 2017, WebNLG 2020, and DART datasets. For the baselines, we fine-tuned LLMs on other graph-to-text generation datasets. We compared three classic datasets: EVENT (EventNarrative) \cite{colas2022eventnarrative}, GN (GraphNarrative) \cite{shi-etal-2023-hallucination}, and TEKGEN \cite{agarwal-etal-2021-knowledge}.  In our training phases, we use the commonly employed cross-entropy loss for generation tasks to align the model's predictions with the grounding generation, which is our PlanGTG dataset. In the full-shot experiments, after fine-tuning on PlanGTG, we continued to fine-tune the models on the training sets of WebNLG 2017. For more details on our training experiments, please refer to the App.\ \ref{ft_ex}}.
\begin{table*}[!ht]
\scriptsize
    \centering
    \begin{tabular}{l|llll|llll|llll}
    \toprule
\textbf{Dataset}&\multicolumn{4}{c}{WebNLG17}&\multicolumn{4}{c}{WebNLG20}&\multicolumn{4}{c}{DART}  \\ 
\cmidrule(lr){2-5}\cmidrule(lr){6-9}\cmidrule(lr){10-13}

        \textbf{\#Metrics}&B-4& ME&CF&BS&B-4& ME&CF&BS&B-4& ME&CF&BS \\
        \midrule
        PlanGTG & 21.13 & 28.69 & 48.96 & -2.41 & 17.62 & 23.20 & 39.04 & -2.60 & 20.45 & 29.37 & 47.10 & -2.82  \\ 
        +reorder & 10.24 & 25.95 & 42.71 & -2.72 & 9.45 & 21.35 & 35.19 & -2.75 & 9.45 & 26.64 & 43.17 & -2.98  \\ 
        +attribution & 18.41 & 26.85 & 46.18 & -2.54 & 16.10 & 23.87 & 38.86 & -2.71 & 15.84 & 27.08 & 44.48 & -2.89  \\ 
        +both (ours) & \textbf{28.76} & \textbf{30.88} & \textbf{50.80} & \textbf{-2.38} & \textbf{20.60} & \textbf{25.50} & \textbf{42.16} & \textbf{-2.51} & \textbf{21.44} & \textbf{31.58} & \textbf{51.30} & \textbf{-2.72} \\ \bottomrule
    \end{tabular}
\setlength{\abovecaptionskip}{0.2cm}
\setlength{\belowcaptionskip}{-0.4cm}
\caption{Ablation study for different modules on
 WebNLG17, WebNLG20 and DART.}
\label{ablation}
\end{table*}
\subsection{ Model Performance}

\noindent \textbf{Zero-shot}:
\vi{Table \ref{6.2.1} demonstrates that integrating the two introduced subtasks significantly enhances our fine-tuned model compared to the baseline LLaMA2-7b-chat and Mistral-7b-chat models. Specifically: 1) Compared to the untuned LLaMA2-7b-chat model, we achieve an average increase of 5.99 points in the BLEU metric and an average increase of 0.3 points in the BARTScore metric; 2) For the other automatically extracted datasets, we observe that their performance is even worse than that of the untuned LLMs. This suggests that  domain differences between the extracted datasets and the downstream tasks are magnified by LLMs. The datasets previously suitable for conventional model pretraining may not translate effectively to LLMs \cite{gardent-etal-2017-webnlg, li-etal-2020-leveraging-large, nan-etal-2021-dart}. Thus, fine-tuning LLMs on PlanGTG demonstrates enhanced domain generalization capabilities.
}


\noindent \textbf{Full-shot}:
To validate the benefit of additional fine-tuning models with PlanGTG on downstream tasks, we conducted additional fine-tuning on WebNLG 2017 for 5 epochs using the checkpoint from the last epoch for testing. This fine-tuning employed the LLaMA2-7b-chat model. For baselines, in addition to the pre-training datasets mentioned in the zero-shot section, we also fine-tuned LLMs using the WebNLG 2017 training dataset, both with and without instructions, following the methodology in our zero/few-shot experiments.

Results presented in Table \ref{full_ft} reveal a surprising decrease in performance when instructions are added compared to fine-tuning alone. This might be attributed to the model's utilization of instructions during inference, leading to outputs that deviate from the standard graph-to-text task format. However, our method, which incorporates instruction tuning, exhibits significant improvements over all baselines. Regarding results on both seen and unseen data, our method outperforms the best baseline (Direct FT) by 0.57 BLEU and 0.23 BARTScore points on seen data, and by 1.64 BLEU and 0.29 BARTScore points on unseen datasets. Similar trends could be observed across the METEOR and CHRF++ metrics, indicating that our method can effectively enhance the model's generalizability capabilities on downstream tasks.

\begin{table*}[!ht]
\scriptsize
    \centering
    \begin{tabular}{l|cccc|cccc|cccc}
    \toprule
\textbf{Dataset}&\multicolumn{4}{c}{WebNLG17}&\multicolumn{4}{c}{WebNLG20}&\multicolumn{4}{c}{DART}  \\
\cmidrule(lr){2-5}\cmidrule(lr){6-9}\cmidrule(lr){10-13}
        \textbf{Methods}&B-4& ME&CF&BS&B-4& ME&CF&BS&B-4& ME&CF&BS \\
\midrule
        One-pass & 23.85 & 34.44 & 55.67 & -2.16 & 25.61 & \textbf{30.97} & 47.60 & -2.34 & 21.67 & 31.00 & 51.41 & -2.54  \\ 
        Baby-steps & \textbf{28.47} & \textbf{38.00} & \textbf{57.84} & -2.39 & \textbf{27.82} & 30.60 & \textbf{49.96} & -2.61 & \textbf{26.99} & \textbf{33.30} & \textbf{54.60} & -2.81  \\ 
        Annealing & 25.40 & 34.09 & 55.16 & \textbf{-2.10} & 27.00 & 30.09 & 49.60 & \textbf{-2.33} & 25.17 & 32.79 & 53.29 & \textbf{-2.51}  \\ 
        Ours & 21.13 & 28.69 & 48.96 & -2.41 & 17.62 & 23.20 & 39.04 & -2.6 & 20.45 & 29.37 & 47.10 & -2.82  \\ \bottomrule
    \end{tabular}
\setlength{\abovecaptionskip}{0.2cm}
\caption{Results achieved by fine-tuning the LLaMA2-7b-chat model on the PlanGTG dataset, without incorporating reordering and attribution, using various curriculum methods.}
\label{curr}
\end{table*}

\begin{table}[]
\setlength{\tabcolsep}{1ex}
\scriptsize
    \centering
    \begin{tabular}{p{20pt}|p{35pt}p{20pt}p{35pt}p{25pt}|p{20pt}} 
    \toprule
    Dataset&Hallucinated Entities&Missed Entities&Hallucinated Relations&Missed Relations&Grammar  \\ \midrule
Gold&0.56&0.08&0.71&0.07&4.58\\ 
LC&2.81&1.95&1.81&1.77&3.09\\
Ours&1.63&1.27&1.49&1.12&3.88\\ \bottomrule
    \end{tabular}
    \setlength{\abovecaptionskip}{0.2cm}
    \setlength{\belowcaptionskip}{-0.4cm}
    \caption{Human evaluation of sentences generated from
   our model trained with PlanGTG and LLaMA2-7b (LC).}
    \label{generated}
\end{table}


\subsection{Ablation Study}
\label{abla}
To further analyze the impact of the two proposed subtasks on model performance, we fine-tuned the LLaMA2-7b-chat model on each subtask and evaluated its zero-shot performance. 

From  results in Table \ref{ablation}, we observe that the performance of PlanGTG with instruction fine-tuning surpasses that of the baseline in Table \ref{6.2.1}. This highlights the effectiveness of PlanGTG even without incorporating the two subtasks. Moreover, incorporating these subtasks leads to additional improvements in model performance, indicating their beneficial impact on enhancing task understanding and output quality.

However, when evaluating the subtasks individually (+reorder and +attribution), the results are less favorable. Introducing the reorder task alone may introduce noise as the model may not fully grasp the significance of the sequence of numbers. Similarly, introducing the attribution task alone could cause a significant mismatch between the model-generated text and the sequence of triplets, thereby degrading text quality.



\subsection{Human Analysis}
\inlineSubsection{Human Evaluation} 
\vi{Following the standards in Section \ref{human_eval}, we conducted a human evaluation involving two assessments of the generated text. We randomly selected 200 graph-to-text pairs from the WebNLG17 dataset, paired with generated texts from the LLaMA2-7b-chat model and LLaMA2-7b-chat model trained with PlanGTG. }

\vi{Results  in Table \ref{generated} show that our model produces texts with higher fidelity and fluency compared to baseline models. In our approach, the reordering subtask improves alignment between the model-generated text and the knowledge graph, while the attribution subtask enhances the model's interpretability of its generated text.}

\subsection{Impact of Curriculum Learning}
\vi{It has been observed that the selection of demonstrations plays a significant role and the difficulty of training examples has an impact on the model results during instruction fine-tuning \cite{lee2024instruction}.  
However, it is unknown whether this also applies to graph-to-text tasks. Given that knowledge graphs contain structured information, we used the number of triplets in a KG as a measure of complexity. 
We leveraged curriculum learning to analyze the impact of progressing from simple to complex learning on the model performance in the graph-to-text instruction fine-tuning process, using PlanGTG without the two subtasks for training. Specifically, we tested three classic curriculum algorithms: (1) \textbf{One-pass} \cite{10.1145/1553374.1553380}, (2) \textbf{Baby-steps}, \cite{spitkovsky-etal-2010-baby} (3) \textbf{Annealing}  \cite{xu-etal-2020-curriculum}, more details are shown in App.\ \ref{ap_curr}. The rest of the training settings are consistent with the experiments in Sec.\ \ref{abla}.}


\vi{Results in  Table \ref{curr} show that compared to instruction fine-tuning without using curriculum learning, all three curriculum learning methods enhance model performance. Among them, Baby-steps obtains  the best results, followed by Annealing, with One-step being the least effective. This suggests that the model may gradually forget the learning of simple graphs (e.g, with 1 or 2 triplets) for text alignment during the learning process from simple to complex. This could potentially impair the model's ability to generate simple sentences, thereby impacting its overall performance.}

\section{Conclusion}
{We have conducted  a comprehensive analysis over both zero- and few-shot scenarios in graph-to-text generation tasks to evaluate the capabilities and challenges of LLMs.
Our findings reveal that (1) LLMs  struggle to understand complex graphs and (2) moderately difficult and diverse demonstrations may help LLMs for translating graphs to text. 
Based on these findings, we propose solutions from both few-shot learning and fine-tuning perspectives to enhance the effectiveness of LLMs. In terms of few-shot prompting, we propose the DDD method to select samples considering both  difficulty and diversity simultaneously, leading to improvements in both the one-shot and few-shot cases. For model fine-tuning, we construct a high-quality  graph-to-text dataset, PlanGTG, and develop two new subtasks. Fine-tuning LLMs on PlanGTG demonstrates a significant improvement in the alignment generation and generalization abilities of LLMs. Additionally, 
through learning from simple to complex data, the model's ability to generate text from graphs is further enhanced. Our work lays the groundwork for future research aimed at effectively enabling LLMs to reorganize graph structures and identify sequential information in generated texts.}

\section*{Limitations}
\vi{We identify the following limitations related to our approach and experiments. Firstly, due to computational resource constraints, we do not evaluate larger models, such as  LLaMA2-70b. Moreover, during fine-tuning, we adopt LoRA, a parameter-efficient fine-tuning method, which, compared to full-parameter fine-tuning, may result in some performance trade-offs compared to full-parameter fine-tuning approaches. Additionally, we face limitations in both budget and computational resources, restricting us from scaling up the dataset or conducting fine-tuning on larger datasets. As a result, PlanGTG is not very large in size. Furthermore, given the focus of the paper on exploring graph-to-text generation  tasks in the era of LLMs, we have not extensively investigated the two new subtasks introduced.}

\bibliography{anthology,custom}


\appendix

\section{Datasets}
\label{app:data}
\begin{table}[]
\centering\scriptsize
\setlength{\tabcolsep}{0.5ex}
\begin{tabular}{lccccc}
\toprule
\multirow{2}{*}{Dataset} &
\multirow{2}{*}{\# train} &
\multirow{2}{*}{\# dev} &
\multirow{2}{*}{\# test} &
\multicolumn{2}{c}{Average}\\
\cmidrule(lr){5-6}
&
&
&
&triplets
&Words
\\
\midrule
  DART &
  30,526 &
  2,768 &
  5,097 &
  3.62 &
  20.95 \\
\midrule
  WebNLG17 (seen) &
  \multirow{2}{*}{6,940} &
  \multirow{2}{*}{\phantom{0,}872} &
  \phantom{0,}971 &
  3.02 &
  20.26 \\
  WebNLG17 (unseen) &
  &
  &
  \phantom{0,}891 &
  2.75 &
  19.00\\
\midrule
  WebNLG20 (seen) &
  \multirow{2}{*}{13,211} &
  \multirow{2}{*}{1,666} &
  \phantom{0,}883 &
  3.63 &
  24.36\\
  WebNLG20 (unseen) &
  &
  &
  \phantom{0,}896 &
  2.71 &
  19.64 \\

\bottomrule
\end{tabular}
\caption{Statistics for the graph-to-text datasets.}
\label{dataset_statsitics}
\end{table}
\textbf{WebNLG17} Challenge \cite{gardent2017creating}: A standard graph-to-text dataset, with each instance being composed of a graph from DBpedia and corresponding text annotated by humans. The test set is divided into the seen and unseen partitions respectively. The unseen partition includes 5 categories absent from the training and development sets. 
\begin{table*}[!ht]
\centering\small
\setlength{\tabcolsep}{0.8ex}
\begin{tabular}{lcccccccccccc}
\toprule
\multirow{2}{*}{Method} &
  \multicolumn{4}{c}{WebNLG17\_unseen} &
  \multicolumn{4}{c}{WebNLG20\_unseen} &
  \multicolumn{4}{c}{DART} \\
\cmidrule(lr){2-5}\cmidrule(lr){6-9} \cmidrule(lr){10-13}
& \multicolumn{1}{c}{B-4} &
  \multicolumn{1}{c}{ME} &
  \multicolumn{1}{c}{CF} &
  \multicolumn{1}{c}{BS} &
  \multicolumn{1}{c}{B-4} &
  \multicolumn{1}{c}{ME} &
  \multicolumn{1}{c}{CF} &
  \multicolumn{1}{c}{BS} &
  \multicolumn{1}{c}{B-4} &
  \multicolumn{1}{c}{ME} &
  \multicolumn{1}{c}{CF} &
  \multicolumn{1}{c}{BS} \\
\midrule
  <head>
  & 18.46  & 34.53 & 54.15 & -2.13 
  & 22.87  & 34.74 & 56.10 & -2.03 
  & 11.57  & 29.97 & 45.68 & -2.70\\
  <head></head>
  & 18.64  & 33.81 & 53.37 & -2.17 
  & \textbf{24.34}  & 30.54 & 50.46 & -2.45 
  & 12.86  & 30.54 & 47.26 & -2.68\\
  head | relation
  & \textbf{19.67}  & \textbf{34.99} & \textbf{54.97} & \textbf{-2.10 }
  & 23.32  & \textbf{35.48} & \textbf{57.07} & \textbf{-1.97}
  & \textbf{13.43}  & \textbf{31.09} & \textbf{48.17} & \textbf{-2.60}\\

\bottomrule
\end{tabular}
\setlength{\belowcaptionskip}{-0.2cm}
\caption{Influence of the triplet formulation.}
\label{triplet_expression}
\end{table*}

\noindent \textbf{WebNLG20} Challenge \cite{castro-ferreira-etal-2020-2020}: It includes 10 categories carried over from WebNLG17 and 5 additional new categories that are not present in the 2017 dataset. Furthermore, this edition introduces a brand-new category ``Company''. 

\noindent \textbf{DART} \cite{nan-etal-2021-dart}: A collection of graph-to-text pairs, which have been compiled from multiple sources, such as WebNLG and E2E \cite{dusek-etal-2018-findings}, along with sentences obtained via crowdsourcing and matching tables sourced from WikiSQL \cite{zhong2017seq2sql} and WikiTableQuestions \cite{pasupat-liang-2015-compositional}. We perform the same partition as \cite{zhao-etal-2023-structure}. The statistics of the datasets are shown in Table~\ref{dataset_statsitics}.
\section{Evaluation Metric Details}
\label{appendix:evaluation_metric}
\inlineSubsection{\vi{Human Evaluation}}\label{hv}
\vi{Following \cite{shi-etal-2023-hallucination}, we assessed the quality of various pretraining datasets and the sentences generated by models. Specifically, our evaluation examined if sentences either from the dataset or generated by models introduced facts not included in the corresponding graphs or failed to mention details. Our analysis employed four metrics: the number of \textbf{hallucinated entities} (entities mentioned in the sentence but absent in the graph), \textbf{missed entities} (entities omitted in the sentence but present in the graph),  \textbf{hallucinated relations} (relations mentioned in the sentence but absent in the graph), and  \textbf{missed relations} (relations omitted in the sentence but present in the graph). Besides, we evaluated the  \textbf{grammatical correctness and fluency} of the generated text. This evaluation utilized a 5-point Likert \cite{zis-Likert1932A} scale, ranging from 1-point (indicating "very poor") to 5-points (indicating "highly satisfactory").} We present an annotation interface and a corresponding example in Fig. \ref{fig2} to demonstrate how humans annotate the quality of PlanGTG.
\begin{figure*}
\begin{tcolorbox}[colframe=black!75!white, colback=white, boxrule=0.5mm, arc=2mm, auto outer arc, width=\linewidth]
\scriptsize 
    \section*{Task Description}
    You are asked to annotate the quality of a graph-text pair based on the following criteria:

    \begin{itemize}
        \item \textbf{Hallucinated Entities}: The number of entities mentioned in the sentence but absent in the graph.
        \item \textbf{Missed Entities}: The number of entities present in the graph but omitted in the sentence.
        \item \textbf{Hallucinated Relations}: The number of relations mentioned in the sentence but absent in the graph.
        \item \textbf{Missed Relations}: The number of relations present in the graph but omitted in the sentence.
        \item \textbf{Grammatical Correctness and Fluency}: Provide a score between 1-5 where:
        \begin{itemize}
            \item 1: Very poor
            \item 2: Poor
            \item 3: Neutral
            \item 4: Fluent
            \item 5: Very natural
        \end{itemize}
    \end{itemize}

    \section*{Graph}
 Indonesia | citytown | Bogor (1) Bogor | population | 950,334 (2) Bogor | altitude | 200 meters (3)

    \section*{Text}
    Located in Indonesia, Bogor is a city with a population of 950,334 and an altitude of 200 meters.

    \section*{Annotation Information}

    \begin{itemize}
        \item \textbf{The number of hallucinated entities:} \underline{\hspace{3cm}}
        \item \textbf{Missed entities:} \underline{\hspace{3cm}}
        \item \textbf{Hallucinated relations:} \underline{\hspace{3cm}}
        \item \textbf{Missed relations:} \underline{\hspace{3cm}}
        \item \textbf{Grammatical correctness and fluency:} \underline{\hspace{3cm}} (1-5)
    \end{itemize}
\end{tcolorbox}
\caption{Human evaluation task annotation interface for the PlanGTG dataset.}
\label{fig2}
\end{figure*}

\smallskip \inlineSubsection{Automatic Evaluation}
\vi{We used four common automatic metrics to assess  graph-to-text  generation: BLEU (B-4) \cite{papineni-etal-2002-bleu}, METEOR (ME) \cite{banerjee-lavie-2005-meteor}, CHRF++ (CF) \cite{popovic-2015-chrf} and BartScore (BS) \cite{yuan2021bartscore}. Specifically, BLEU measures the $n$-gram overlaps between the generated text and  reference text. We set $n$ to 4. CHRF++ computes the F-score averaged on both character and word-level $n$-grams. METEOR considers the semantic matches between source and reference text and BartScore uses BART \cite{lewis-etal-2020-bart} to measure the quality of the generated text. }

\section{Preliminary Evaluation}
\label{appendix:evaluation}

\subsection{Prompt Searching}

\paragraph{\vi{Instruction Format Exploration}}
\label{appendix:instruction_explore}

\begin{table*}[]
\centering\small
\setlength{\tabcolsep}{1ex}
\begin{tabular}{llcccccccccccc}
\toprule
\multirow{1}{*}{SYS} &
\multirow{1}{*}{USR} &
  \multicolumn{4}{c}{WebNLG20\_all} & \multicolumn{4}{c}{WebNLG17\_all} & \multicolumn{4}{c}{DART} \\
\cmidrule(lr){3-6}\cmidrule(lr){7-10}\cmidrule(lr){11-14}
  Prompt& 
  Prompt &  
  \multicolumn{1}{c}{B-4} &
  \multicolumn{1}{c}{ME} &
  \multicolumn{1}{c}{CF} &
  \multicolumn{1}{c}{BS} &  
  \multicolumn{1}{c}{B-4} &
  \multicolumn{1}{c}{ME} &
  \multicolumn{1}{c}{CF} &
  \multicolumn{1}{c}{BS} &
  \multicolumn{1}{c}{B-4} &
  \multicolumn{1}{c}{ME} &
  \multicolumn{1}{c}{CF} &
  \multicolumn{1}{c}{BS} \\
\midrule
\multicolumn{1}{l}{\multirow{5}{*}{Simple}} &
  A &
 \phantom{0}8.97 & 24.12 & 42.59 & -2.70 & \phantom{0}9.03 & 23.90 & 41.94 & -2.81 & \phantom{0}5.57 & 21.79 & 36.92 & -3.29\\
 &
  B &
10.80 & 25.09 & 43.86 & -2.58 & \phantom{0}9.35 & 22.40 & 41.84 & -2.91 & \phantom{0}5.77 & 21.32 & 35.80 & -3.41\\
 &
  C &
10.03 & 24.76 & 42.53 & -2.67 &  \phantom{0}9.68 & 24.75 & 42.27 & -2.66 & \phantom{0}5.58 & 22.08 & 36.91 & -3.17\\
 &
  D &
\textit{24.22} & \textit{34.65} & \textit{58.07} & -2.07 &\textbf{22.72} & \textit{34.65} & \textit{53.12} & \textbf{-2.09} & \textbf{15.26} & \textit{30.71} & \textit{47.87} & \textit{-2.62}\\
 &
  D* &
21.66 & 33.84 & 54.89 & -2.10 & 18.12 & 32.53 & 51.78 & -2.22 & \textit{13.44} & 30.52 & 47.33 & -2.68 \\
\midrule
\multicolumn{1}{l}{\multirow{5}{*}{Detailed}} &
  A &
  15.28 & 31.64 & 49.70 & -2.10& 13.77 &  31.29 & 47.87 & -2.33 & 10.12 & 28.32 & 43.12 & -2.79\\
 &
  B &
  14.33 & 31.07 & 48.70 & -2.21 & 14.37 &  31.68 & 48.64 & -2.32 & 10.49 & 28.60 & 43.79 & -2.81\\
 &
  C &
  20.88 & 34.27 & 55.74 & \textit{-2.03} & 17.69 &  33.59 & 52.34 & -2.17 & 12.38 & 29.75 & 46.22 & -2.70\\
 &
  D &
  24.00 & \textbf{36.13} & \textbf{58.17} & \textbf{-1.91} & \textit{19.39} &  \textbf{34.93} & \textbf{54.95} & \textbf{-2.09} & 13.43 & \textbf{31.00} & \textbf{48.04} & \textbf{-2.61}\\
 &
  D* &
  \textbf{24.29} & 29.69 & 49.17 & -2.61 & 18.06 & 33.64 & 43.00 & -2.85 & 11.03 & 26.42 & 40.70 & -3.09\\

\bottomrule
\end{tabular}
\setlength{\belowcaptionskip}{-0.2cm}
\caption{Zero-shot LLaMA2 performance between different prompts. The best results are marked \textbf{bold} and the second best results are marked \textit{italics}}
\label{inst_explore}
\end{table*}
\vi{Since prompt is the key to the interaction between the LLMs and humans, we started by investigating how the verbalization of instructions influences the downstream performance. For system prompts, we constructed a simple version where only a general instruction is given and a detailed version where the definition of triplets and  a detailed instruction are provided. For the user prompt, we constructed four instructions sorted by their level of detail in the prompts from A to D. In addition,  to simulate the controllable text generation in industrial applications, we constructed D*, where an additional task of writing triplet dollar signs after generating the text is added. Table~\ref{inst_explore} presents the performance results for WebNLG20 and WebNLG. The prompt templates used are shown as follows}
\begin{tcolorbox}[colback=gray!20]
\textbf{Simple}: ``Following the questions and give directly the answers. Do not include any additional information or outputs.''

\textbf{Detailed}:``You are skilled in interpreting knowledge graphs. Your task is to transform a series of triplets, each consisting of a subject, predicate, and object, into a well-written, coherent paragraph. These triplets are formatted as 'subject | predicate | object' and are separated by lines. Please provide only the transformed text as your output. Do not include any additional information or outputs.''
\end{tcolorbox}
\begin{tcolorbox}[colback=gray!20]

\textbf{A}:``\{triplets\} || Text: ''

\textbf{B}:``Graph: \{triplets\} || Text: ''

\textbf{C}:``Graph: \{triplets\} || Convert the graph into text: ''

\textbf{D}:``Following is a set of knowledge graph triplets delimited by triplet backticks, each on a separate line, in the format: subject | predicate | object.
  \\`\\`
  \{triplets\}
  \\`\\`
  Only use information from the provided triplets and convert the graph into a coherent piece of text:''

\textbf{D*}:``Following is a set of knowledge graph triplets delimited by triplet backticks, each on a separate line, in the format: subject | predicate | object.
  \\`\\`
  \{triplets\}
  \\`\\`
  Only use information from the provided triplets and generate a coherent piece of text that contains all of the information in the triplets.
  After you finish writing the piece of text, write triplet dollar signs (i.e.: \$\$\$).''
\end{tcolorbox}

\vi{From the results in Table~\ref{inst_explore}, we can draw the following conclusions:  (1) When the user prompt is less detailed, giving detailed system prompts instead helps the quality of the generation, especially for the semantics level  since a significant improvement is observed on CHRF++ and Bart score. This suggests that a detailed description of the task may hint the model to generate more fluent and human-preferred answers. (2) More detailed user prompts may result in better generations as stable improvements for all metrics are obtained for prompts of  type A to D. We also observe a sudden growth in the performance for both  system settings for types C to  D in all datasets.  (3) Including a controlled text generation task (user prompt D*) may harm the quality of the generation since a significant drop in all metrics is observed. This is possibly due to the weaker capabilities of smaller LLMs such as LLaMA-7b, where multiple tasks may interfere with each other. Based on the above findings, we therefore choose the prompt settings with the best performance (a detailed system prompt with the template D and the third triplet format) as the input in the other experiments.}

\paragraph{triplet representation}\vi{We also want to explore the influence of different verbalizations of triplets within the graphs. To this end, we experiment with three popular verbalizations of  triplets.   Table \ref{triplet_expression} presents the results, where a close performance of each verbalization is observed. This suggests that LLMs have a good understanding of  triplet expressions.}

\label{appendix:linearizaion}
\begin{table}[]
\centering\scriptsize
\setlength{\tabcolsep}{0.8ex}
\begin{tabular}{lccccc}
\toprule
  \multicolumn{1}{c}{Dataset} &
  \multicolumn{1}{c}{Linearization} &
  \multicolumn{1}{c}{B-4} &
  \multicolumn{1}{c}{ME} &
  \multicolumn{1}{c}{CF} &
  \multicolumn{1}{c}{BS} \\
\midrule
  \multirow{5}{*}{WebNLG17\_all} &
  RS  & 28.39 & 28.61 & 52.21 & -2.56 \\
&ORI & 28.55 & 28.70 & 52.15 & \textbf{-2.54} \\
&BFS & 28.61 & \textbf{28.74} & \textbf{52.37} & \textbf{-2.54} \\
&DFS & 28.59 & 28.65 & 52.20 & \textbf{-2.54} \\
&GPT & \textbf{28.89} & 28.34 & 51.64 & -2.56 \\

  \midrule
  \multirow{5}{*}{WebNLG20\_all} &
  RS  & 31.64 & 30.22 & 54.93 & -2.42 \\
  &ORI & 32.37 & 30.60 & 55.60 & -2.37 \\
  &BFS & \textbf{32.44} & 30.64 & 55.62 & -2.36 \\
  &DFS & 32.34 & 30.55 & 55.47 & -2.37 \\
  &GPT & 30.58 & \textbf{32.13} & \textbf{57.66} & \textbf{-2.26} \\
  \midrule
  \multirow{5}{*}{DART} &
  RS  & \textbf{22.45} & \textbf{27.83} & 51.79 & -2.99 \\
  &ORI & 20.32 & 27.69 & 51.72 & -3.01 \\
  &BFS & 20.35 & 27.72 & \textbf{51.80} & \textbf{-3.02} \\
  &DFS & 20.32 & 27.70 & 51.76 & \textbf{-3.02}\\
  &GPT & 20.93 & 27.40 & 51.04 & \textbf{-3.02} \\

\bottomrule
\end{tabular}
\caption{Comparison between different linearizations}
\label{Linearizaion}
\end{table}
\paragraph{Influence of Linearization}
\vi{Massive works have demonstrated the impact of various linearization techniques on graph-to-text tasks \cite{yang-etal-2020-improving-text, hoyle-etal-2021-promoting, li-etal-2021-shot-knowledge}. Here we investigated how different linearizations may affect the results of LLMs. Beyond the original linearization (ORI) provided in a dataset, we explored alternative arrangements by reordering the triplets according to fixed tree traversal methods, including breadth-first search (BFS), depth-first search (DFS), and a random sequence (RS). Additionally, we employed GPT to generate a `silver' linearization derived from the ground truth text, detailed prompts can be found in the Fig \ref{linearizaion}. The results of experiments performed on the Mistral-7b-instruct-v0.2 are shown in Table \ref{Linearizaion}. They reveal marginal differences among the linearizations, with BFS and DFS exhibiting slightly enhanced performance. The discrepancy between random and silver linearizations is also minimal, indicating that LLMs demonstrate robustness to the variety of input graph linearizations. Consequently, we opt for the ORI linearization for subsequent experiments in our study.}

\subsection{Few-shot Sample Selection}
\label{appendix:example_selection}
Few-shot demonstrations have widely been shown to be crucial on the performance of generations \cite{luo2023dr, drozdov2023parade}. We conducted experiments  on how different example selection strategies influence the performance of the graph-to-text generation, focusing on criteria such as difficulty and diversity. Experiment results inspire us to propose an optimal strategy for demonstration selection.
\subsubsection{Demonstration  Selection Methods}
\vi{To measure the diversity and difficulty, we first map the graph into continuous vectors using a state-of-the-art sentence encoder\footnote{We used the SoTA sentence embedding encoder UAE \cite{li2023angle} on the MTEB leader board.} and then use the embeddings to achieve  the difficulty and diversity-based demonstration selection. } 
\begin{table*}[!thp]
\centering\small
\setlength{\tabcolsep}{0.8ex}
\begin{tabular}{lcccccccccccc}
\toprule
& 
\multicolumn{4}{c}{WebNLG17\_all} & 
\multicolumn{4}{c}{WebNLG20\_all} &   
\multicolumn{4}{c}{DART}\\
\cmidrule(lr){2-5}\cmidrule(lr){6-9}\cmidrule(lr){10-13}
Models &          B-4 &         ME &         CF &         BS &          B-4 &         ME &         CF &         BS &    B-4 &          ME &          CF &          BS \\
\midrule
\multicolumn{13}{c}{\textbf{Instruct-Models}} \\
ChatGLM3-6b &        12.69 &     26.28 &     45.27 &      -2.53 &        17.16 &     29.03 &     49.75 &      -2.34 &  17.21 &      30.61 &      50.27 &        -2.30 \\
Vicuna-7b-v1.5 &        13.89 &      30.70 &     47.45 &      -2.34 &         17.20 &     30.86 &     49.04 &      -2.21 &  14.23 &      31.03 &      47.87 &       -2.36 \\
Zephyr-7b-beta &        17.05 &     34.74 &     53.74 &      -2.08 &        21.51 &     36.01 &     56.83 &      -1.95 &  12.27 &       31.50 &      47.49 &       -2.28 \\
Falcon-7b-instruct &        13.77 &     26.65 &     43.87 &      -2.96 &        15.35 &     26.79 &     44.42 &      -2.86 &  10.95 &       22.10 &      38.06 &       -3.33 \\
LLaMA2-7b-chat-hf &        21.51 &     35.67 &     56.38 &      -2.04 &        24.68 &     35.84 &     57.82 &      -1.96 &  15.47 &      31.94 &       49.70 &       -2.27 \\
Gemma-1.1-7b-it &        19.77 &     30.30 &     50.40 &      -2.72 &        18.10 &     27.08 &     45.48 &      -3.01 &  12.36 &      25.56 &       42.33 &       -3.09 \\
Mistral-7b-Instruct-v0.2 &        \textbf{27.09} &     27.29 &     45.43 &      -2.73 &        \textbf{31.12} &     29.35 &     49.12 &      -2.54 &  21.48 &      24.22 &        41.00 &       -3.07 \\
LLaMA2-13b-chat-hf &        22.51 &     \underline{36.31} &     \underline{57.62} &      \underline{-1.95} &         27.70 &     \underline{37.63} &     \underline{60.47} &      \underline{-1.83} &  17.47 &      \underline{33.37} &      {52.04} &       \underline{-2.14} \\
Vicuna-13b-v1.5 &        25.19 &     35.24 &     56.95 &      -2.06 &        \underline{30.82} &      35.60 &     57.88 &      -1.91 &  \underline{22.42} &      32.35 &      51.96 &       -2.28 \\
Falcon-40b-instruct &        21.02 &     34.17 &     55.03 &       -2.20 &         26.80 &      35.40 &       58.00 &      -2.07 &  20.78 &      33.31 &      \underline{53.68} &       -2.25 \\
Mixtral-\(\text{8}\times\text{7b}\) &        \underline{26.42} &     \textbf{38.06} &      \textbf{60.70} &      \textbf{-1.84} &        30.49 &     \textbf{38.41} &     \textbf{62.07} &      \textbf{-1.78} &  \textbf{22.89} &      \textbf{35.42} &      \textbf{56.04} &          \textbf{-2.00} \\
\midrule
\multicolumn{13}{c}{\textbf{Base-Models}} \\
Falcon-7b-base &   \phantom{0}3.82 &     15.93 &     26.17 &      -2.26 &    \phantom{0}3.48     &     15.15 &     26.03 &      -2.36 &  \phantom{0}3.36 &  16.27 &  25.16 &  -2.44 \\
Gemma-7b &   \phantom{0}0.08 &  \phantom{0}1.73 &    \phantom{0}3.93 &      -5.25 &    \phantom{0}0.05     &    \phantom{0}1.44 &    \phantom{0}3.56 &   -5.33 &  \phantom{0}0.09 & \phantom{0}1.57 &  \phantom{0}3.85 &  -5.18 \\
LLaMA2-7b-hf &   \phantom{0}4.57 &    16.61 &    27.94 &      -2.83 &    \phantom{0}4.42     &  16.13 &     28.64 &      -2.98 &  \phantom{0}3.98 &  16.87 &  26.37 &  -2.82 \\
LLaMA2-13b-hf &   \phantom{0}3.83 &    14.57 &    26.35 &      -3.02 &    \phantom{0}4.20     &   15.11 &     27.51 &      -3.15 &  \phantom{0}3.25 &  14.52 &  24.25 &  -3.11 \\
\bottomrule
\end{tabular}
\setlength{\abovecaptionskip}{0.3cm}
\setlength{\belowcaptionskip}{-0.4cm}
\caption{Graph-to-text generation performance of  the tested LLMs. The best results are  bold and the second best results are underlined. }
\label{scaling_results}
\end{table*}
\begin{table}[!th]
\centering\scriptsize
\begin{tabular}{llccccc}
\toprule
\multirow{1}{*}{Dataset} &
  \multicolumn{1}{c}{Method} &
  \multicolumn{1}{c}{Shot} &
  \multicolumn{1}{c}{BLE} &
  \multicolumn{1}{c}{MET} &
  \multicolumn{1}{c}{CHR+} &
  \multicolumn{1}{c}{BRT} \\
\midrule
  \multirow{4}{*}{DART} &
  Difficulty & 1 & 19.30 & 33.14 & 54.01 & -2.39 \\
  &DDD       & 1 & \textbf{21.33} & {33.69} & {55.10} & {-2.36} \\
  &Diversity & 3 & 21.31 & 33.71 & 55.00 & -2.35 \\
  &DDD       & 3 & {21.30} & \textbf{33.78} & \textbf{55.17} & \textbf{-2.33} \\
  \midrule
  \multirow{4}{*}{WebNLG17} &
  Difficulty & 1 & 22.46 & 35.27 & 57.20 & -1.96 \\
  &DDD       & 1 & {23.48} & {35.49} & {57.66} & {-1.92} \\
  &Diversity & 3 & \textbf{25.19} & 35.61 & {58.30} & -1.93 \\
  &DDD       & 3 & {25.07} & \textbf{35.98} & \textbf{58.60} & \textbf{-1.91} \\
  \midrule
  \multirow{4}{*}{WebNLG20} &
  Difficulty & 1 & 24.92 & 35.81 & 57.76 & -1.98 \\
  &DDD       & 1 & {26.07} & \textbf{38.01} & {58.64} & \textbf{-1.97} \\
  &Diversity & 3 & 26.65 & 35.92 & 58.32 & -2.02 \\
  &DDD       & 3 & \textbf{27.54} & {36.09} & \textbf{58.93} & {-2.00} \\
\bottomrule
\end{tabular}
\setlength{\abovecaptionskip}{0.2cm}
\setlength{\belowcaptionskip}{-0.2cm}
\caption{Comparison between different samplings on DART, WebNLG17 unseen and WebNLG20 unseen datasets.}
\label{DDD}
\end{table}

We assess difficulty by computing the cosine similarity between examples and input. It is based on the assumption that following similar demonstrations facilitates graph-to-text generation. Consistent patterns in similar demonstrations enable LLMs to produce coherent text that fits the norms of the demonstrated examples. We categorize difficulty into five levels based on cosine similarity scores between input embeddings and all embeddings in the training sets. For experimentation, we uniformly select examples ranging from the easiest (Difficulty level 0) to the most challenging (Difficulty level 4) and conducted one-shot inference.

Inspired by \citet{liu-etal-2022-makes}, we measure the diversity of examples based on  different $k$-means clusters and conducted a 3-shot experiment. We introduced a four-tier diversity-based sampling framework designed to enhance example selection. Initially, we identify the least diverse examples by selecting the nearest $n$ points to the input embeddings (Level 0). For Level 1, we sample $n$ points within the same cluster as the input. At Level 2, we select the centers of the $n$ closest clusters to the input's cluster. Finally, Level 3 involves a uniform selection of cluster centers based on their proximity to the input, specifically choosing the centers of the first, fifth, and tenth nearest clusters from a total of ten in our evaluation.  

For each strategy, different verbalization of demonstrations are also considered. we calculate the similarity or $k$-means not only between graphs of the demonstration-input pairs, but also between the text of the input obtained by a zero-shot inference. Furthermore, a standardized graph, where entities are replaced with ``anonymous'' entities $\langle \textit{ent} \rangle$, is  constructed to study whether it is the semantic information in the graph or its topology that primarily influences graph verbalization.

\subsubsection{\vi{Difficulty-based Sampling}}  
\vi{From the results in Figure~\ref{Difficulty_full}, we observe that: (1) Sampling by text similarity shows good performance in the BLEU metric across all samples. This suggests that it could be a good strategy when prioritizing $n$-gram accuracy and being provided additional inference for zero-shot text from input graphs is feasible; (2) There is no significant difference on the performance of standardized graphs and original graphs, suggesting that LLMs learn more from the structure of graphs than from the semantic information in entities; (3) When comparing the difficulties between selected demonstrations, we observe that while an easy demonstration works better on the datasets  where the triplets  have appeared in the training sets (e.g.\ DART), this trend does not hold for datasets with unseen samples in the test set (e.g., WebNLG). Instead, optimal performance typically peaks with samples of moderate difficulty (levels 1-3), indicating that selecting moderately challenging samples may be the optimal strategy}


\subsubsection{\vi{Diversity-based Sampling}}
From Figure \ref{Diversity_full}, two main findings are: Firstly, text embedding-based demonstrations generally outperform others, indicating that diversity has limited impact when using text similarity as the selecting criteria. This suggests that zero-shot learning  
offers a more straightforward and efficient method for accessing detailed information than relying on graph embeddings. Secondly, the selection of examples based on graph embeddings underscores the significance of diversity. Contrary to the lowest diversity level (level 0), optimal performance is typically achieved at moderate diversity levels (levels 1 or 2). This emphasizes the effectiveness of a balanced approach to select examples, favoring samples from nearby clusters rather than those within the same cluster or from distant clusters.

\subsubsection{DDD Selection}
The full comparison results of DDD selection against difficulty-based selection and diversity-based selection are shown in Table \ref{DDD}.


\subsection{\vi{Scaling Evaluations}}
\label{sec:scaling_evaluaions}
We evaluate popular LLMs with the chosen prompt settings (necessary modifications are made to adapt the templates for different models) and default configs from the HuggingFace text generation pipeline for experiments. We set the maximum token limit for each model to ensure that we obtain complete results. For models around 7 billion parameters, we test ChatGLM-6b \cite{zeng2023glm-130b}, LLaMA2-7b \cite{touvron2023llama}, Vicuna-7b \cite{vicuna2023}, Zephyr-7b \cite{tunstall2023zephyr}, Mistral-7b \cite{jiang2023mistral} and Falcon-7b \cite{falcon40b}.  For larger LLMs, we test, falcon-40b and  Mixtral-45b (the official 8*7b mixture of experts version).The results are presented in Table \ref{scaling_results}. We also evaluate 4 base models. However, since we found that all base models suffer from following instructions and only generate the reference, we draw our conclusion mainly based on instruct models.

\begin{table}[]
\centering\small
\setlength{\tabcolsep}{0.6ex}
\begin{tabular}{lccccc}
\toprule
  \multicolumn{1}{c}{Dataset} &
  \multicolumn{1}{c}{\#triplets (\#samples)} &
  \multicolumn{1}{c}{B-4} &
  \multicolumn{1}{c}{ME} &
  \multicolumn{1}{c}{CF} &
  \multicolumn{1}{c}{BS} \\
  \midrule
  \multirow{5}{*}{DART} &
  1\phantom{00} (848) & 23.36 & 38.93 & 61.22 & -1.92 \\ 
&2\phantom{00} (797) & 17.55 & 32.98 & 53.59 & -2.36 \\
&3\phantom{00} (821) & 17.09 & 32.57 & 53.24 & -2.22 \\
&4\phantom{00} (869) & 17.21 & 31.68 & 52.89 & -2.29 \\
&5+ (1762) & 11.79 & 29.59 & 47.08 & -2.63 \\
  \midrule
  \multirow{5}{*}{WebNLG17} &
  1\phantom{0} (454) & 30.73 & 41.65 & 70.16 & -1.61 \\ 
&2\phantom{0} (349) & 29.82 & 30.93 & 55.14 & -2.63 \\
&3\phantom{0} (386) & 26.54 & 27.41 & 50.51 & -2.97 \\
&4\phantom{0} (363) & 25.07 & 24.99 & 46.75 & -3.05 \\
&5+ (310) & 19.87 & 21.71 & 41.34 & -3.32 \\
  \midrule
  \multirow{5}{*}{WebNLG20} &
  1\phantom{0} (369) & 30.64 & 41.51 & 69.18 & -1.66 \\ 
&2\phantom{0} (349) & 32.76 & 32.72 & 57.10 & -2.44 \\
&3\phantom{0} (350) & 30.92 & 30.18 & 54.51 & -2.60 \\
&4\phantom{0} (305) & 28.61 & 26.93 & 49.81 & -2.79 \\
&5+ (406) & 28.88 & 26.94 & 51.38 & -2.85 \\
\bottomrule
\end{tabular}
\setlength{\belowcaptionskip}{-0.1cm}
\caption{Comparison between different graph triplet numbers on DART, WebNLG17\_all, WebNLG20\_all.}
\label{graph_complex}
\end{table}
Our key findings are as follows. Firstly, 13b models outperform almost all 7b models, indicating a positive correlation between the scaling of LLMs and the graph-to-text generation performance. However, the increasing trend becomes slower when we keep scaling up the parameters. We hypothesize that this is because 13b models are already capable of understanding and reasoning well over graphs, scaling parameters may help LLMs to memorize more facts, but this is not helpful for graph-to-text verbalization. Additionally, we observe that a fine-tuned version of LLMs usually performs worse than the vanilla ones, as seen from Zephyr-7b to Mistral-7b and Vicuna-7b to LLaMA2-7b. This suggests that task-specific fine-tuning may compromise the graph verbalization capabilities of LLMs, which further motivates the design of tasks aimed at instruction-tuning LLMs to enhance their graph-to-text capabilities.

\subsection{Influence of Graph Complexity}
\label{Appendix: GraphComplexity}

\vi{We also investigated the impact of graph complexity on the performance of graph-to-text generation by LLMs. Conducted in a zero-shot framework using Mistral-7b-instruct-v0.2, our experiments in Table \ref{graph_complex} reveal a consistent decline in performance across all metrics as the number of triplets in a graph increases. This shows the difficulties that LLMs face in text generation from complex graphs. Additionally, we examine the effect of graph diameters (the longest-shortest path between any two vertices) on LLM performance. The results presented in Table \ref{graph_diameter} show that for WebNLG20 the variance in performance across different graph diameters is minimal, potentially suggesting model stability across various structures. However, the results for  WebNLG17 exhibit a clear negative correlation between performance metrics and graph diameters, highlighting the current limitations of LLMs in handling diverse graph structures. This inconsistency underscores the necessity for further research to enhance LLM stability across different graph configurations.}
\begin{figure}[t]
\centering
\includegraphics[width=1\linewidth,height=0.6\linewidth]{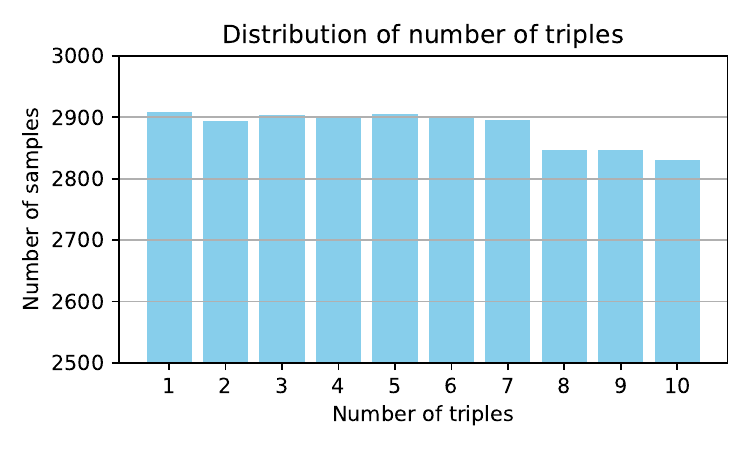}
\setlength{\belowcaptionskip}{-0.4cm}
\setlength{\abovecaptionskip}{-0.4cm}
\caption{Bar chart of the distribution of the number of triplets in PlanGTG}
\label{PlanGTG_graph_triplet}
\end{figure}

\begin{figure}[t]
\centering
\includegraphics[width=1\linewidth,height=0.6\linewidth]{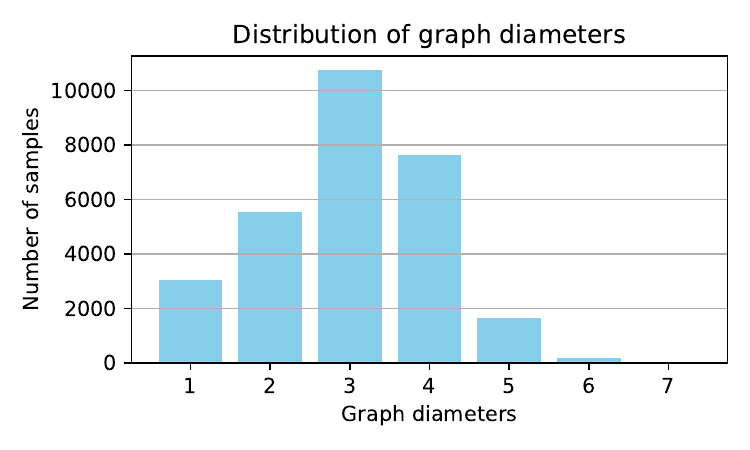}
\setlength{\abovecaptionskip}{-0.4cm}
\setlength{\belowcaptionskip}{-0.4cm}
\caption{Bar chart of the distribution of the diameter of triplets in PlanGTG}
\label{PlanGTG_graph_dia}
\end{figure}

\begin{figure}[!t]
\centering
\includegraphics[width=1\linewidth,height=0.6\linewidth]{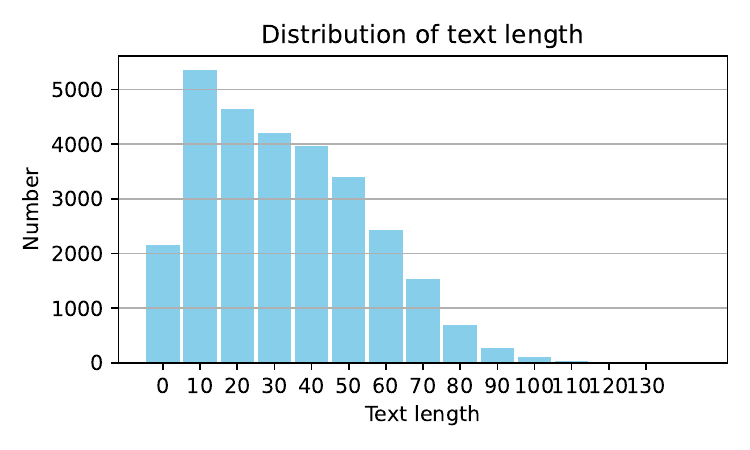}
\setlength{\abovecaptionskip}{-0.4cm}
\setlength{\belowcaptionskip}{-0.4cm}
\caption{Bar chart of the distribution of the length of text in PlanGTG}
\label{PlanGTG_text}
\end{figure}
\begin{table}[]
\centering\small
\setlength{\tabcolsep}{0.6ex}
\begin{tabular}{lccccc}
\toprule
  \multicolumn{1}{c}{Dataset} &
  \multicolumn{1}{c}{\#D (\#samples)} &
  \multicolumn{1}{c}{B-4} &
  \multicolumn{1}{c}{ME} &
  \multicolumn{1}{c}{CF} &
  \multicolumn{1}{c}{BS} \\
  \midrule
  \multirow{5}{*}{DART} &
  0\phantom{0} (575)\phantom{00} & 23.29 & 38.92 & 61.19 & -1.93 \\
&1\phantom{0} (3463)\phantom{0} & 14.18 & 30.30 & 48.65 & -2.53 \\
&2\phantom{0} (850)\phantom{00} & 18.60 & 33.20 & 55.09 & -2.00 \\
&3\phantom{0} (201)\phantom{00} & 17.75 & 32.37 & 54.41 & -2.02 \\
&4\phantom{0} (8)\phantom{0000} & 20.57 & 33.96 & 59.36 & -2.24 \\
  \midrule
  \multirow{5}{*}{WebNLG17} &
  0\phantom{0} (454)\phantom{0} & 30.73 & 41.65 & 70.16 & -1.61 \\
&1\phantom{0} (930)\phantom{0} & 27.92 & 27.57 & 50.58 & -2.84 \\
&2\phantom{0} (337)\phantom{0} & 22.07 & 23.37 & 44.03 & -3.18 \\
&3\phantom{0} (136)\phantom{0} & 16.66 & 20.23 & 38.81 & -3.44 \\
&4\phantom{0} (5)\phantom{0}\phantom{0}\phantom{0} & 12.38 & 17.99 & 38.17 & -3.64 \\
  \midrule
  \multirow{5}{*}{WebNLG20} &
  0\phantom{0} (369)\phantom{0} & 30.77 &  41.42 &  69.08 & -1.66 \\
&1\phantom{0} (1042) & 29.84 &  28.47 &  51.86 & -2.63 \\
&2\phantom{0} (280)\phantom{0} & 30.60 &  27.71 &  52.66 & -2.88 \\
&3\phantom{0} (88)\phantom{0}\phantom{0} & 30.80 &  29.29 &  55.57 & -2.64 \\
\bottomrule
\end{tabular}
\setlength{\belowcaptionskip}{-0.4cm}
\caption{Comparison between graph diameters (D).}
\label{graph_diameter}
\end{table}
\begin{figure*}[t]
\centering
\includegraphics[width=1\linewidth,height=0.33\linewidth]{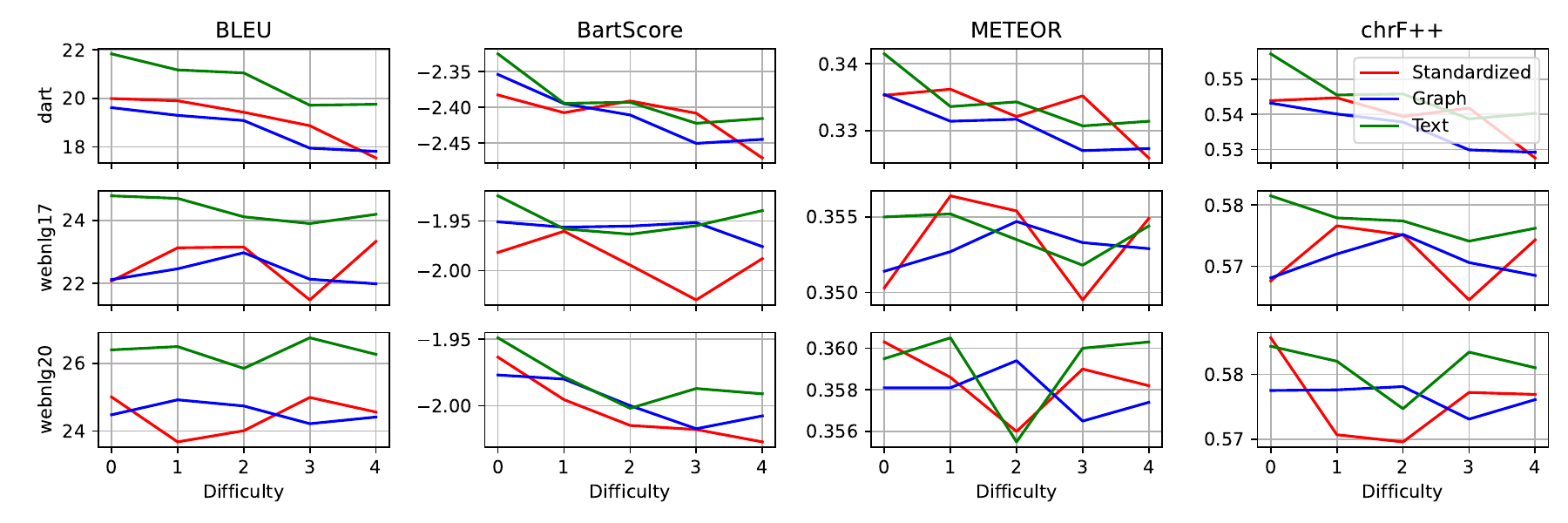}
\caption{Results for all 4 metrics exploring the difficulty of examples in the one-shot learning setting}
\label{Difficulty_full}
\end{figure*}
\begin{figure*}[t]
\centering
\includegraphics[width=1\linewidth,height=0.33\linewidth]{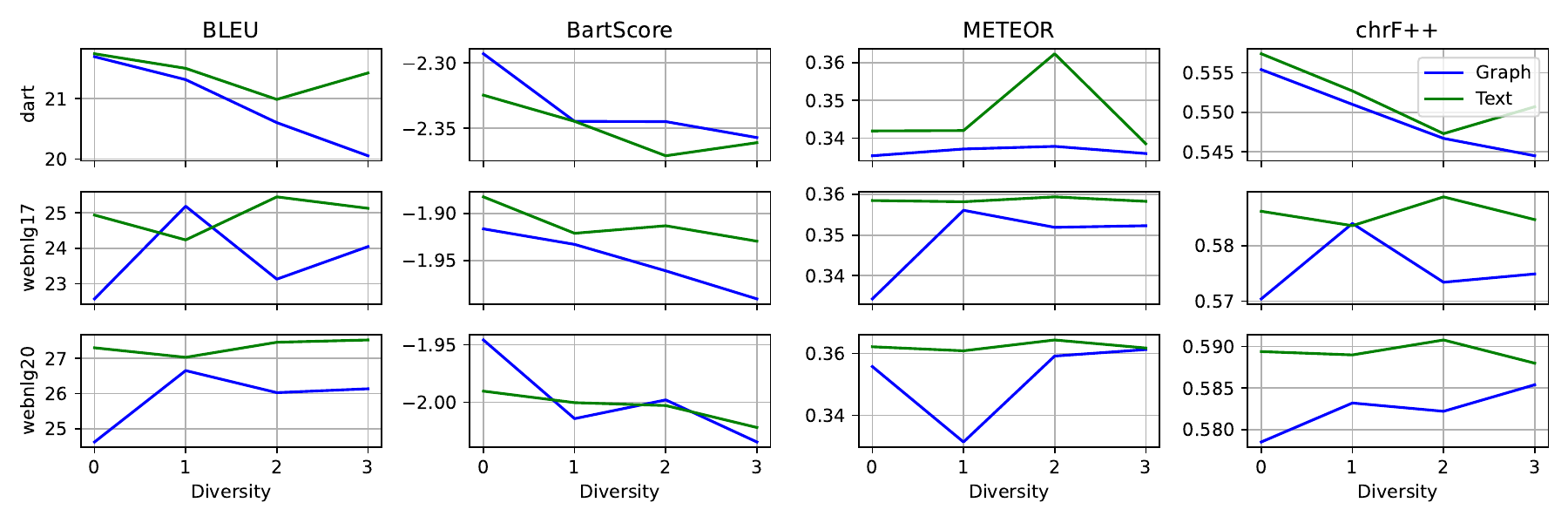}
\caption{Results for all 4 metrics exploring the diversity of examples in the few-shot learning setting}
\label{Diversity_full}
\end{figure*}


\section{PlanGTG Details}
\subsection{Details for prompts in PlanGTG}
\begin{figure*}[t]
\centering
\includegraphics[width=1\linewidth,height=1.3\linewidth]{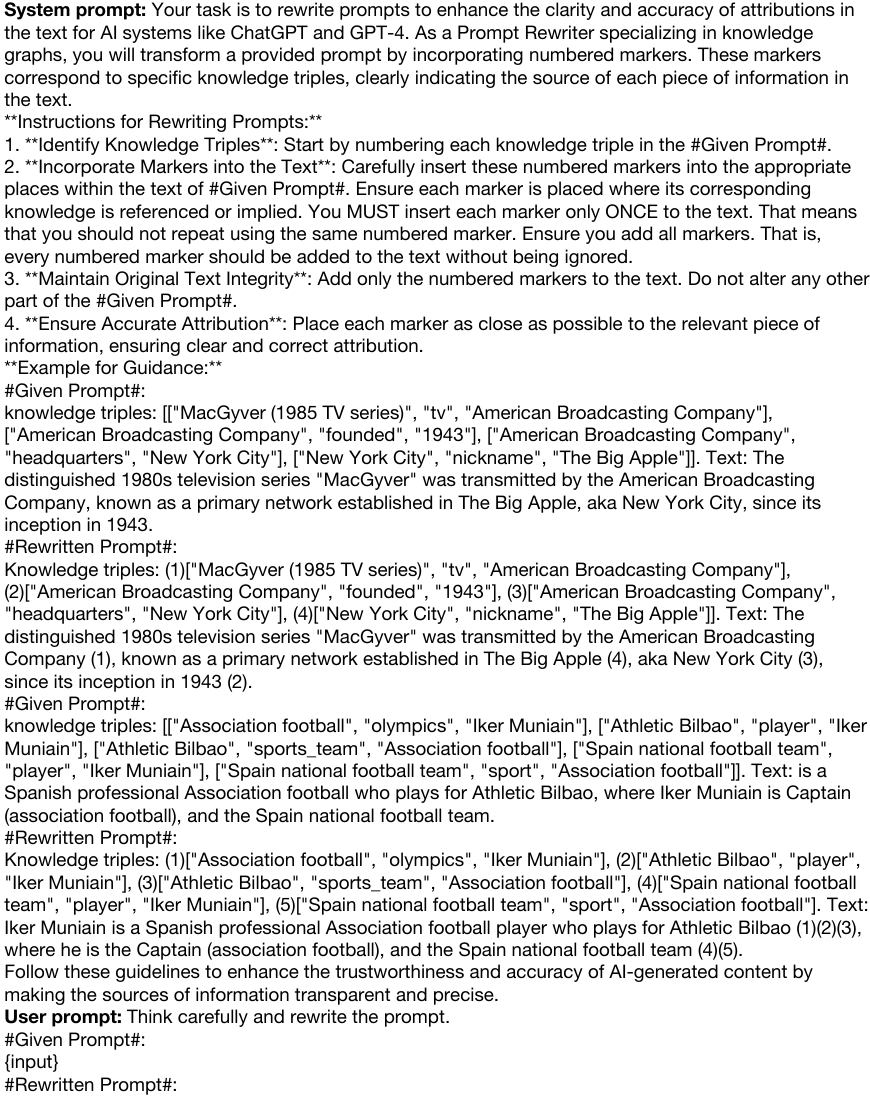}
\caption{Prompt for the parallel attribution annotation}
\label{PlanGTG_prompt_parallel}
\end{figure*}

\begin{figure*}[t]
\centering
\includegraphics[width=1\linewidth,height=0.55\linewidth]{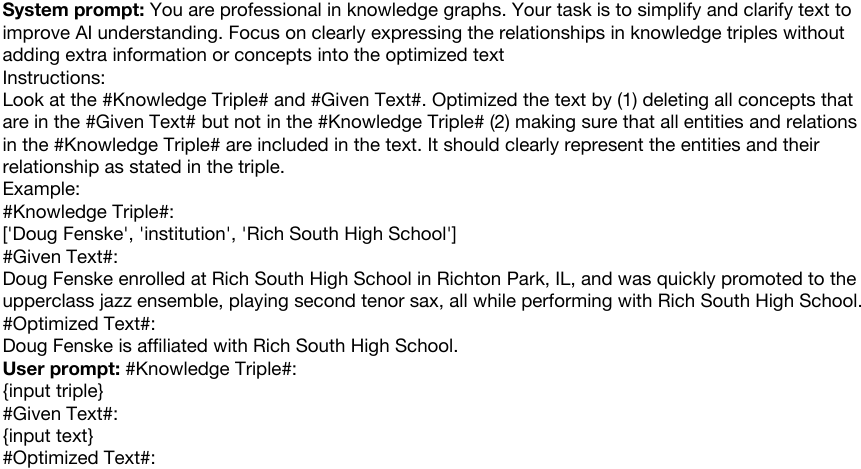}
\caption{Prompt for the rewrite and annotation}
\label{PlanGTG_prompt_rewrite}
\end{figure*}

\begin{figure*}[t]
\centering
\includegraphics[width=1\linewidth,height=1.15\linewidth]{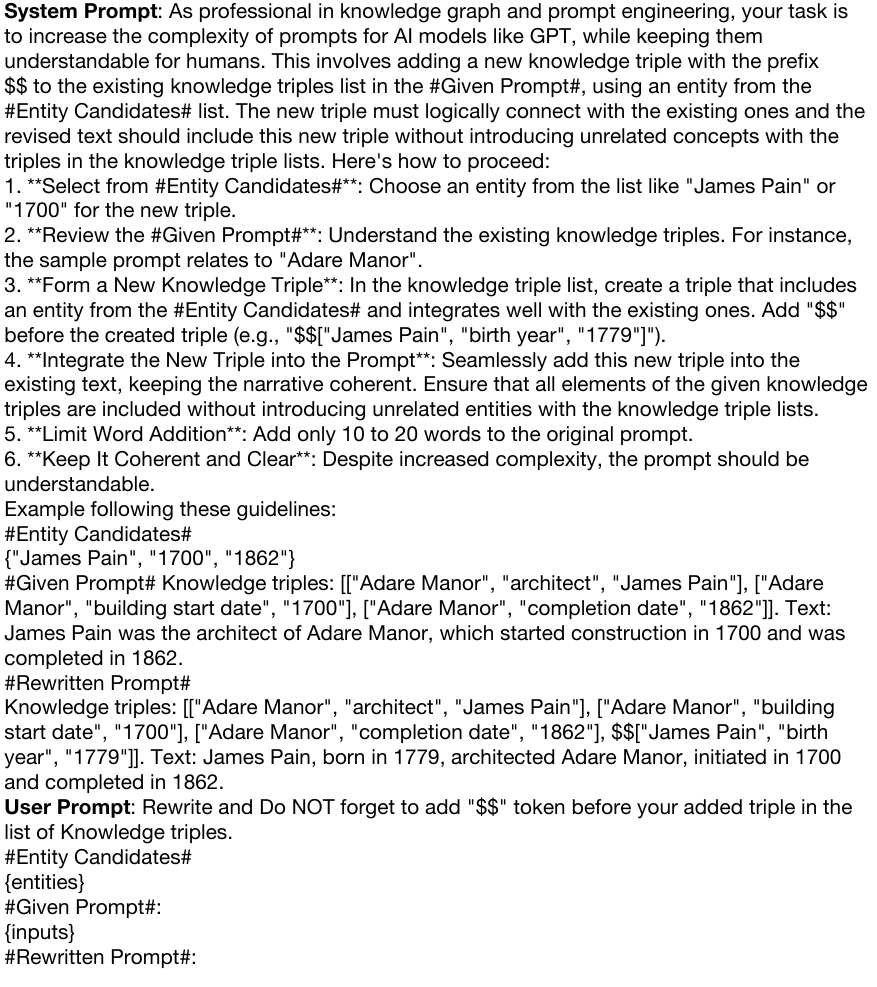}
\caption{Prompt for the sequential generation, where the \{entities\} and \{inputs\} are the given input}
\label{PlanGTG_prompt_sequential}
\end{figure*}

\begin{figure*}[t]
\centering
\includegraphics[width=1\linewidth,height=0.7\linewidth]{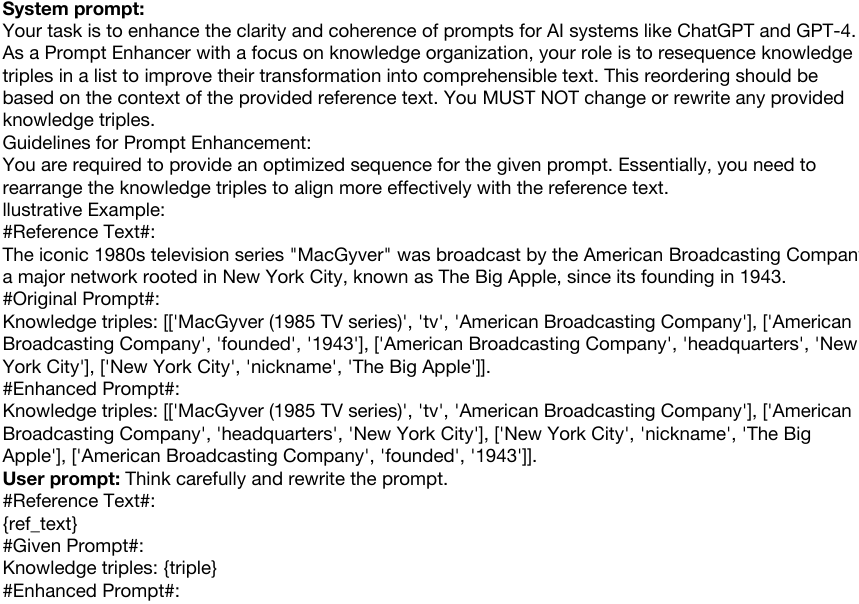}
\caption{Prompt for the linearization annotation}
\label{linearizaion}
\end{figure*}

The used prompts are shown in Fig. \ref{PlanGTG_prompt_rewrite}, Fig. \ref{PlanGTG_prompt_sequential}, Fig. \ref{PlanGTG_prompt_parallel} respectively.

\subsection{Details for the distribution in PlanGTG}
\label{Appendix: PlanGTGDescription}
The distribution of number of triplets, graph diameters and text length of PlanGTG are shown in Fig. \ref{PlanGTG_graph_triplet}, Fig. \ref{PlanGTG_graph_dia} and Fig. \ref{PlanGTG_text} respectively.

\section{Details for our training experiments}
We mainly used two popular LLMs to conduct the experiments: LLaMA2-7b-chat\footnote{https://huggingface.co/meta-llama/Llama-2-7b-chat-hf} and Mistral-7b-chat\footnote{https://huggingface.co/mistralai/Mistral-7B-Instruct-v0.2}. To optimize memory usage and accelerate training, we applied DeepSpeed Zero Stage 3 \cite{10.1145/3394486.3406703} and bfloat16 mixed precision techniques. The learning rate was set at 2e-4 for all experiments with a batch size of 8, and the maximum length was 1024. All models were trained on 2 Tesla A100-80G GPUs.

When fine-tuning on our PlanGTG dataset, given that we involve two tasks, namely reorder and attribution, our instructions are set as follows:
\begin{tcolorbox}[colback=gray!20]
The following is a set of knowledge graph triplets delimited by triplet backticks, each on a separate line, in the format: subject | predicate | object. (number). The ``number'' indicates the sequence number of each triplet.
``
    {triplets}
''
The task involves two steps: First, output the correct order of these knowledge triplets. Then, generate a coherent piece of text that incorporates all the information from the triplets. The generated text should include corresponding sequence numbers. Only the information provided in the triplets should be used. After you finish these two tasks, write triplet dollar signs (i.e.: \$\$\$).
\end{tcolorbox}
The input to the model follows the format: subject | predicate | object (number). And The model's gold output format is: The correct triplet order is: subject | predicate | object (number). The generated text is: ... . Below is an example:
\begin{tcolorbox}[colback=gray!20]
\noindent \textbf{Input:}

Twitter | users | 330 million (1)
Twitter | founding year | 2006 (2)
Twitter | category | Internet forum (3)
Internet forum | communication platform | online discussion platform (4)

\noindent \textbf{Output:} \\
\textbf{The correct triplet order is}: 
Twitter | users | 330 million (1)
Internet forum | communication platform | online discussion platform (4)
Twitter | category | Internet forum (3)
Twitter | founding year | 2006 (2) \textbf{Then The generated text is}: Twitter has 330 million users (1), serving as an online communication platform for discussions (4), is categorized as an internet forum (3), and was founded in 2006 (2).
\end{tcolorbox}

We finetune the model on our constructed PlanGTG instruction dataset, enabling the model to automatically learn to first output the correct order of the graph, and then generate text based on the reordered triplets. During inference, we extract the text following ``Then the generated text is:'' and remove the indices (e.g., (1)) before evaluating against the gold text. The fine-tuning loss we adopt is the commonly used cross-entropy loss for generation tasks, which measures the difference between the predicted probability distribution and the true distribution (i.e., the gold text). Specifically, the cross-entropy loss is defined as:

\[
L = - \sum_{i=1}^{N} \sum_{j=1}^{V} y_{ij} \log(p_{ij})
\]

where \(N\) is the sequence length, \(V\) is the vocabulary size, \(y_{ij}\) is the one-hot encoded vector for the gold token at position \(i\), and \(p_{ij}\) is the predicted probability for token \(j\) at position \(i\). The model is trained to minimize this loss, encouraging it to generate text that closely matches the gold output.

When fine-tuning on the WebNLG17 dataset, our instructions are set as follows because this dataset only provides annotations for graphs and their corresponding text:
\begin{tcolorbox}[colback=gray!20]
The following is a set of knowledge graph triplets delimited by triplet backticks, each on a separate line, in the format: subject | predicate | object.
``
\{triplets\}
''
Generate a coherent piece of text that contains all of the information in the triplets. Only use information from the provided triplets. After you finish writing the piece of text, write triplet dollar signs (i.e.: \$\$\$).
\end{tcolorbox}

\label{ft_ex}

\section{Case study}
Our model's superior performance is attributed to its ability to better align generated knowledge graph sequences with the target text, as demonstrated in Table \ref{case}. This alignment is achieved through the sequence reordering task, which guides  the model to produce sequences that closely match the intended narrative structure, such as reordering from (1)$\rightarrow$(2)$\rightarrow$(3) to (3)$\rightarrow$(2)$\rightarrow$(1). This task also enables accurate annotation of sequence numbers within the generated text, enhancing coherence and fidelity.

In contrast, the outputs from the model without fine-tuning fail in both aspects, and their sequence annotations are disordered. Furthermore, these outputs sometimes include hallucinated information. In the first example, the second and third triplets do not exist, yet the model generates two unrelated and even hallucinated triplets, thus introducing errors into the generated text.

\label{case_study}
\begin{table*}
\scriptsize
        \centering
    \begin{tabular}
    {p{120pt}|p{175pt}|p{112pt}}
    \toprule
      \textbf{Input instruction and gold reference} & \textbf{Sentence generated by LLaMA2-7b-chat} & \textbf{Sentence generated by our trained model} \\ \midrule 
    {\textbf{Input:} (Texas, LargestCity, Houston) \circled{1}
    
    \textbf{Reference:} Houston is the largest city in Texas}
&\textbf{Order:} (Texas, LargestCity, Houston) \circled{1}, \textcolor{violet}{(Texas, capital, Austin) \circled{2}, (Houston, LargestCity, Texas) \circled{3}} 
  \xmark 

\textbf{Text:} In the state of Texas, the largest city is Houston, \textcolor{purple}{which is also the capital of Texas.} \circled{1},  \textcolor{violet}{\circled{2}, \circled{3}} \xmark 
&\textbf{Order:} (Texas, LargestCity, Houston) \circled{1}, 

\textbf{Text:} Houston is the largest city in Texas \circled{1}. \cmark
\\ \midrule
{\textbf{Input:} (Albert Jennings Fountain, DeathPlace, New Mexico Territory) \circled{1}, (Albert Jennings Fountain, BirthPlace, New York City) \circled{2}, (Albert Jennings Fountain, BirthPlace, Staten Island) \circled{3}

\textbf{Reference:} Albert Jennings Fountain was born in Staten Island, New York City and died in the New Mexico Territory.}
&\textbf{Order:} (Albert Jennings Fountain, DeathPlace, New Mexico Territory) \circled{0}, (Albert Jennings Fountain, BirthPlace, New York City) \circled{1}, (Albert Jennings Fountain, BirthPlace, Staten Island) \circled{2}

\textbf{Text:} Albert Jennings Fountain was born in New York City \circled{1}, and later moved to New Mexico Territory \circled{0} where he passed away. \textcolor{purple}{According to the knowledge triplets}, he was born on Staten Island \circled{2} \xmark
&\textbf{Order:} (Albert Jennings Fountain, BirthPlace, Staten Island) \circled{3} (Albert Jennings Fountain, BirthPlace, New York City) \circled{2}, (Albert Jennings Fountain, deathPlace, New Mexico Territory) \circled{1} 

\textbf{Text:} Albert Jennings Fountain, born in Staten Island \circled{3}, New York City \circled{2}, died in the New Mexico Territory \circled{1}. \cmark
\\ \bottomrule
\end{tabular}
\setlength{\abovecaptionskip}{0.2cm}
\setlength{\belowcaptionskip}{-0.4cm}
\caption{ Two sample texts generated by LLaMA2-7b-chat baseline and our trained model. The wrong outputs are marked red and the hallucinated outputs are marked purple. The first example shows that LLMs suffer less from hallucination after being tuned by the attribution sub-task. The second example shows that LLMs can better understand relations between triplets and generate more fluent text after unlocking the planning capability.}
\label{case}
\vspace{-0.2cm}
\end{table*}
\section{Three Curriculum Algorithms}
\label{ap_curr}
1) \textbf{One-pass algorithm} \cite{10.1145/1553374.1553380}: The training data \(D\) is sorted by graph difficulty and distributed into \(k\) number of buckets. We train the model from the simplest buckets to the most complex.

\noindent 2) \textbf{Baby-steps curriculum} \cite{spitkovsky-etal-2010-baby}: which first distributes the sorted data into buckets (or shards/bins) from easy to hard and starts training with the easiest bucket. After a fixed number of training epochs or convergence, the next bucket is merged into the training subset. Finally, after all the buckets are merged and used, we also fine-tune on it once. 

\noindent3) \textbf{Annealing scheduler} proposed by \citet{xu-etal-2020-curriculum}: as with (1) and (2), we start training from the easiest bucket, but for the next training bucket, we randomly add 1/k examples from the current and previous bucket. 

\end{document}